\documentclass{article}

     \usepackage[preprint]{neurips_2019}

\usepackage[utf8]{inputenc} 
\usepackage[T1]{fontenc}    
\usepackage{url}            
\usepackage{booktabs}       
\usepackage{amsfonts}       
\usepackage{nicefrac}       
\usepackage{microtype}      

\usepackage{microtype}
\usepackage{graphicx}
\usepackage{hyperref}
\usepackage{subfigure}
\usepackage{bm}
\usepackage{amsmath, amssymb}
\usepackage{mathtools}
\usepackage{bbm}
\usepackage{bm}
\usepackage{booktabs} 
\allowdisplaybreaks

\DeclareMathOperator*{\argmax}{arg\,max}

\usepackage[]{placeins}
\usepackage{chngcntr}
\usepackage{siunitx}
\usepackage{scalerel,stackengine}

\usepackage{algorithm}
\usepackage{customalgorithm2e}

\usepackage{wrapfig}

\title{A General Framework for Abstention Under Label Shift}

\author{%
  Amr M. Alexandari*\\
  Department of Computer Science\\
  Stanford University\\
  \texttt{amr.alexandari@gmail.com} \\
  \And
  Anshul Kundaje$\dagger$\\
  Departments of Genetics \& Computer Science\\
  Stanford University\\
  \texttt{anshul@kundaje.net} \\
  \And
  Avanti Shrikumar*$\dagger$\\
  Department of Computer Science\\
  Stanford University\\
  \texttt{avanti.shrikumar@gmail.com} \\
  \\
  \textbf{*co-first authors}
  \textbf{$\dagger$ co-corresponding authors}
}

\begin{document}

\maketitle

\begin{abstract}
In safety-critical applications of machine learning, it is often important to abstain from making predictions on low confidence examples.  Standard abstention methods tend to be focused on optimizing top-k accuracy, but in many applications, accuracy is not the metric of interest.  Further, label shift (a shift in class proportions between training time and prediction time) is ubiquitous in practical settings, and existing abstention methods do not handle label shift well.  In this work, we present a general framework for abstention that can be applied to optimize any metric of interest, that is adaptable to label shift at test time, and that works out-of-the-box with any classifier that can be calibrated.  Our approach leverages recent reports that calibrated probability estimates can be used as a proxy for the true class labels, thereby allowing us to estimate the change in an arbitrary metric if an example were abstained on.  We present computationally efficient algorithms under our framework to optimize sensitivity at a target specificity, auROC, and the weighted Cohen's Kappa, and introduce a novel strong baseline based on JS divergence from prior class probabilities.  Experiments on synthetic, biological, and clinical data support our findings.
\end{abstract}

\section{Introduction}

Abstention, or selective classification, is a setting in which models can flag a subset of difficult or low-confidence cases for a human expert instead of making predictions. The goal of abstention is to optimize a metric of interest or to guarantee a minimum level of performance by the system. Abstention approaches have been proposed to safeguard automated diagnostic systems so that a medical diagnosis model would not classify with high confidence when it should be flagging difficult cases for human intervention \citep{medical}. More recently, there has been a great effort to automate certain diagnostic tasks so that healthcare workers are free to perform more pressing tasks. As a result, several diagnostic systems have been proposed with an abstention option in place. For example, \citet{Leibig2017-ie} proposed a convolutional neural network model for the detection of sight-threatening diabetic retinopathy, and performed abstention based on Monte-Carlo dropout \citep{testtime} in order to surpass the NHS Diabetes Eye Screening guidelines of 85\% sensitivity and 80\% specificity. However, Monte-Carlo dropout is not tailored to optimize sensitivity at a target specificity, which is the clinical target metric.

\begin{figure}[!ht]
\centerline{\includegraphics[width=0.6\textwidth]{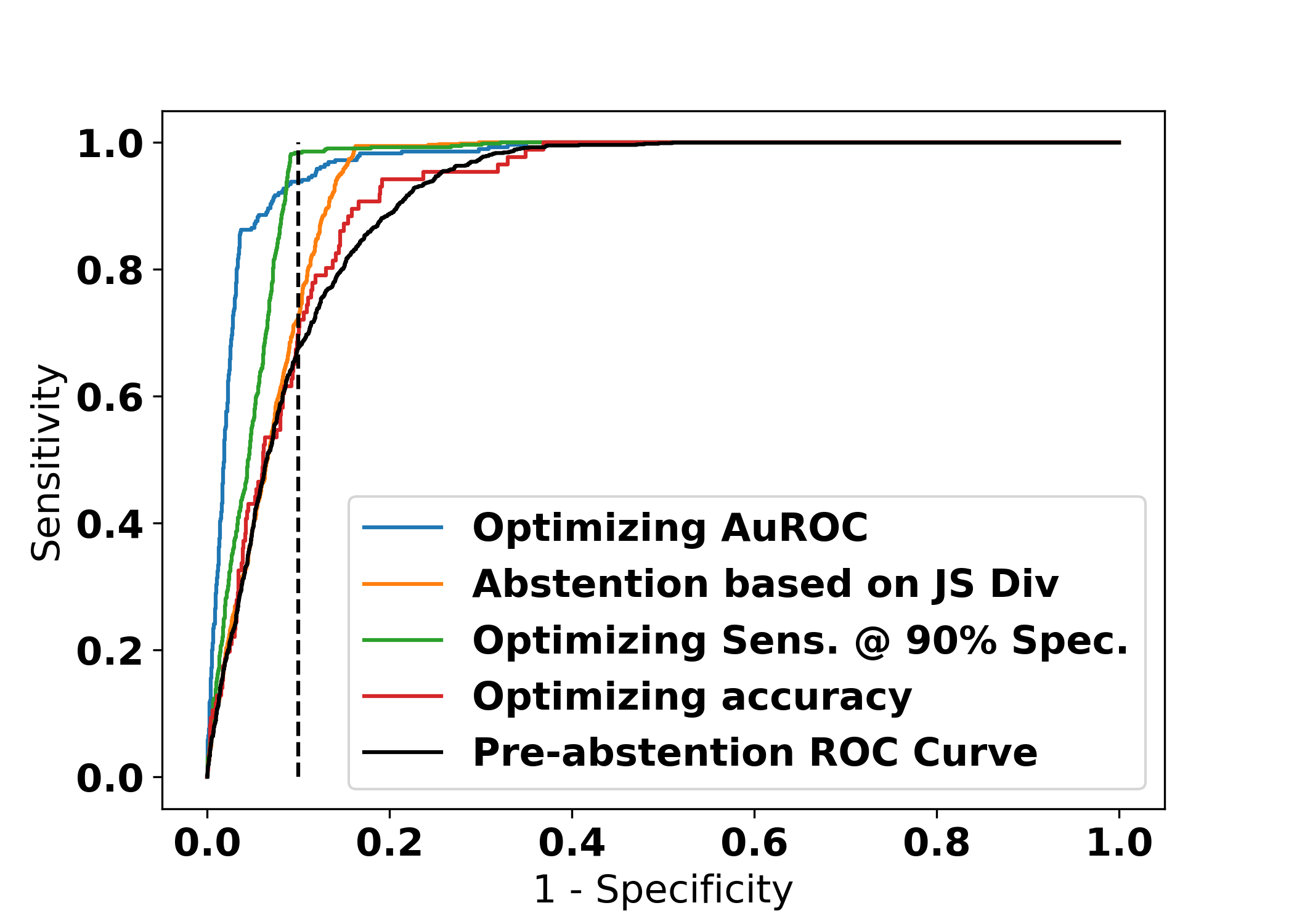}}
\caption{\small \textbf{ Optimal abstention boundaries can vary dramatically depending on the metric of interest}. Shown are ROC curves before and after abstention for a simulated binary classification dataset where the ratio of positives:negatives was 1:9. 30\% of examples were abstained on using three approaches introduced in this work: abstaining to optimize the AuROC (blue line),  abstention based on JS divergence introduced in Sec. \ref{sec:js-div} (orange line), abstaining to optimize the sensitvity at a specificity of 90\% (green line), and abstention to optimize accuracy (red line). The solid black line represents the ROC curve prior to abstention, and the dashed black line is the 90\% specificity threshold. We can see that the metric-specific abstention algorithms perform the best at optimizing their respective target metrics. See Sec. ~\ref{sec:simulatedbinarydataset} for details of the simulation.}
\label{fig:importanceofmetricspecific}
\end{figure}

In fact, to our knowledge, no abstention approach has been proposed specifically to optimize any metric other than accuracy. Yet, accuracy has shortcomings (e.g. predicting majority class achieves high accuracy when class proportions are highly unbalanced), and in practice metrics other than accuracy are used - for example, recall at a specific FDR, the area under the ROC Curve, or Cohen's Kappa. Unlike top-k accuracy or weighted misclassification error, these performance metrics depend on the overall distribution of predictions rather than errors on individual predictions. This makes them challenging to use with approaches that require the cost of abstention and/or misclassification to be specified on a per-example basis \citep{Cordella1995-qy,De_Stefano2000-ub,Pietraszek2005-df, selectiveclassification}. To illustrate this issue, consider the simple case where we abstain on a binary classifier's predictions according to the distance of the predicted probability from 0.5, which corresponds to abstaining based on the probability of the most confident class (the classes being `positive' or `negative'). We will assume that the classifier's predicted probabilities have been calibrated, as this is achievable in practice using a held-out dataset such as a validation set \citep{calibration}. The choice of 0.5 as a threshold might be motivated by the notion that 0.5 is the prediction that the model makes in the absence of other information. While this is true for balanced IID datasets, it is far from true if the label distribution is non-uniform -- for example, if the ratio of positives to negatives is 1:9, a calibrated model would make a prediction of 0.1 in the absence of any other information. We could alter our abstention rule to use 0.1 as the threshold instead of 0.5, but this still fails to consider the specific metric we are attempting to optimize. For example, optimizing the sensitivity at a specificity of 90\% may require abstaining on a very different set of examples compared to optimizing the overall area under the ROC Curve (Fig. ~\ref{fig:importanceofmetricspecific}).

An additional limitation of many existing abstention approaches is that they are not robust to distribution shifts, where the new data distribution differs from the training set distribution. Distribution shifts occur regularly in many practical settings and violate the empirical risk minimization assumption that the data distribution during training is identical to the data distribution at test time, thereby posing a challenge to practitioners whose goal is to ensure the reliability of their system. Label shift is an important form of distribution shift that is particularly pervasive in medical settings. In label shift, the prior class probability $p(y)$ changes between the training and test distributions, while the conditional probability $p(\boldsymbol{x}|y)$ stays fixed. Label shift corresponds with anti-causal learning \citep{Schoelkopf2012-px}, and is different from covariate shift \citep{sugiyama2007covariate} and concept shift \citep{lu2018learning}.

The label shift phenomenon can be understood with an example. Suppose we wish to deploy an automated diagnostic system to classify whether a person has a particular disease from displayed symptoms for the purposes of public health monitoring. Assume the prevalence of this disease in the population $p(y)$ is 0.01\%. We train a classifier that predicts 0.01\% positive on the training data and 0.01\% on withheld data. We deploy the classifier in the clinic and it predicts 0.01\% in the patient population. After a few months, there is an outbreak of the disease and the classifier predicts an overwhelming 1\% positive in the patient population. Note that the nature of the symptoms caused by the disease $p(\boldsymbol{x}|y)$ has not changed with its prevalence $p(y)$. Without accounting for label shift, it is unclear how to interpret this 1\% positive predictions, because the IID assumptions under which the classifier was developed have already been violated. Is the disease prevalence actually 1\% or more than it? Which diagnoses can be trusted and which can not?

Although approaches have been developed to handle label shift \citep{Saerens2002-jh,storkey2009training,Schoelkopf2012-px,lipton2018detecting}, the interface of label shift and abstention remains underexplored. Unfortunately, standard abstention approaches do not handle label shift well. Moreover, \citet{kahneman1, kahneman2} showed that people too are not reliable at mentally evaluating the prior probabilities of outcomes. Interestingly, even when humans update their priors without supervision, they are susceptible to the Test-Item Effect which can be induced under label shift. In particular, humans can classify the same item in opposite ways, depending on what other test items they are asked to classify (without label feedback) \citep{zhu2010cognitive}. In other words, algorithms are needed for this problem.

\section{Our Contributions}

Recently, \citet{biascorrectedtempscaling} showed that a simple maximum likelihood approach, when coupled with bias-correction terms in the calibration formula, is effective at producing well-calibrated conditional probabilities $p(y|x)$ even under label shift. This approach is referred to as MLLS. Follow up work by \citet{garglabelshift} provided consistency conditions for MLLS as well as a decomposition of its finite-sample error into terms reflecting miscalibration and estimation errors. This was also explicitly studied under label shift. As \citet{garglabelshift} noted, while it is true that the calibration in \citep{biascorrectedtempscaling} is not theoretically guaranteed to produce a consistent estimate (for that, an approach explicitly satisfying canonical calibration would be needed - and approaches for canonical calibration do exist), in practice the calibration used in MLLS is good enough to significantly outperform competing approaches for label shift.

In this work, we leverage MLLS for calibration and contribute a \textbf{general framework for abstention} that can be applied to optimize an \textbf{arbitrary metric of interest}. The core insight of our approach is that calibrated predicted probabilities can be used to estimate the change in the metric when particular examples are abstained on. Although intuitive, this approach has not been used before even in situations where it could have improved the results, and abstention approaches that were not tailored to the target metric of interest were used instead \citep{Leibig2017-ie, jones2020selective}. Under our framework, we propose \textbf{computationally efficient algorithms} for optimizing the change in the sensitivity at a particular specificity, the area under the ROC curve, and the weighted Cohen's Kappa. We demonstrate that our framework is \textbf{adaptive to label shift} in that it can generalize to a test-set distribution with an unknown prior probability shift. We compare against several baselines (both pre-existing and ones we introduce here) to demonstrate the effectiveness of our approach.

For the reader's clarity, we note that the focus of this work is not out-of-distribution detection (that is, we are not identifying individual examples that lie outside the training data distribution). Even when all test-set examples for a given class are drawn from the same distribution as the corresponding training set class, abstention could still be needed to meet a performance threshold; that is the setting we are interested in.

\section{Problem Formulation}
\label{sec:problemformulation}

The abstention objectives typically fall into the categories of \emph{bounded} and \emph{cost-based} \citep{Pietraszek2005-df}. In the bounded abstention case, the goal is to achieve the maximum improvement in a performance metric of interest while abstaining on no more than a fraction $k$ of instances, or to abstain on as few examples as possible while attaining a performance equal to or better than some target performance. The two bounded objectives are equivalent in the sense that if one possesses an abstention rule for the former, one can search over different $k$ to obtain an abstention rule for the latter.

In the cost-based abstention scenario, there is a cost associated with the model's performance and also a cost associated with abstaining on examples. The objective here is to minimize the total cost. In the case where the cost is monotonically increasing in the total number of abstained examples and monotonically decreasing in the performance of the model, cost-based abstention is equivalent to bounded abstention in the sense that if one possesses an abstention rule for optimizing performance while abstaining on a fraction $k$ of instances, one can again search over different values of $k$ to obtain an abstention rule for the cost-based case.

In many practical applications of machine learning, the bounded abstention case is the dominant case encountered \citep{fumeramultiplethresholds}. Due to the aforementioned equivalence between cost-based abstention and bounded abstention under assumptions of monotonicity, in this work we explored the bounded abstention case. We report results in terms of the improvement in a metric of interest when abstaining on no more than a fraction $k$ of instances.

\section{Prior Work}
\label{sec:preexistingbaselines}

Many families of solutions have been explored for abstention, yet a general framework that can work for any target metric has not been proposed \citep{Hellman1970-zo, Fumera2002-gh, Bartlett2008-qf, El-Yaniv2010-xs, Cortes2016-cm}. Both \citet{henNgim} and \citet{selectiveclassification} abstained on examples according to the probability of the most confidently predicted class. Formally, if $p(y=i|x)$ represents the predicted probability that example $x$ belongs to class $i$, then this amounts to abstaining on all examples $x$ for which $\max_i p(y=i|x) < t$, where the value of $t$ is selected to satisfy the bounded abstention criterion. In the case of binary classification, this method is equivalent to abstaining on examples based on the distance of the predicted probability from $0.5$. In a different work, \citet{entropyabstention1} abstained on examples that have the highest entropy in their predicted probabilities. For binary classification, this is again equivalent to abstaining based on the distance from $0.5$. Neither method account for the prior class probabilities or the performance metric of interest.

An approach that could be adapted to account for a specific performance metric of interest was proposed in \citet{fumeramultiplethresholds}. Rather than using a single threshold $t$ for all classes (as is done in the abstention rule $\max_i p(y=i|x) < t$), \citet{fumeramultiplethresholds} proposed a rule that abstains on all examples $x$ for which $i^* = \argmax_i p(y=i|x) \wedge p(y=i^*|x) < t_{i^*}$. In \citet{fumeramultiplethresholds}, the class-specific thresholds $t_{i^*}$ were selected by a brute force iterative search to optimize accuracy on the validation set - however, in this work we adapt the method to optimize metrics other than accuracy. In the binary classification case, this amounts to abstaining on examples where the predicted probability $p(y=1|x)$ falls in an asymmetric interval around $0.5$ (that is, $(1-t_0) < 0.5 < t_1$). There are three drawback to this approach. First, because the threshold is selected according to $\argmax_i p(y=i|x)$, the region of abstention necessarily includes the point where all classes are predicted with equal probability. Although it may naively seem that one would always want to include this point in the abstention region, this is only true if the metric we wish to optimize is accuracy. If, instead, we wish to optimize a different metric - say the area under the ROC curve in a binary classification task with a non-uniform prior probability distribution - the optimal abstention region may not include the point with a predicted probability of $0.5$. Second, because thresholds are optimized according to what performs well on the validation set, we cannot expect this approach to generalize to a test-set with an unknown label shift -- in other words, it is not \emph{adaptive}. Finally, a brute force search over abstention thresholds is computationally expensive, particularly when there are multiple classes involved.

Of course, Bayesian methods can be used to obtain uncertainty estimates and abstain on predictions with high uncertainty. In biomedical settings, some Bayesian methods rely on simplified prior models and the resulting inferences can deviate significantly from those indicated by the true posterior \citep{awate2019estimating}. In Bayesian deep learning, the weights of the network are considered to be drawn from a prior distribution rather than having fixed values \citep{Bayesian0:online,bayesianhypernets}. This typically requires special techniques for training the model and cannot be applied retroactively to an existing trained model. One exception is test-time dropout or Monte-Carlo dropout \citep{testtime}, which is based on the observation that leaving dropout enabled during prediction time is equivalent to approximate inference in a Gaussian process model. When using this method in our experiments, we leave dropout enabled at test-time and compute the variance in the output across 100 sampled predictions. For the multi-class case, we take the variance in the probability of the most confidently predicted class, as was done in \citet{selectiveclassification}. The examples that exhibit the highest variance are abstained on.

\section{A General Framework for Abstention under Label Shift}
\label{sec:frameworkoverview}

We first discuss the overall framework, which can accommodate any metric of interest, and which is adaptive to unknown label shift in the test-set. Our adaptability comes from the fact that the method relies only on the calibrated probabilities - thus, any approach that adapts calibrated probabilities to cope with label shift, such as \citet{biascorrectedtempscaling}, naturally applies to this framework without modification (assuming the label shift assumption holds).

On the most basic level, our framework can be summarized as a simple 3-step strategy:  \textbf{(1)}  use a heldout portion of the training set (such as the validation set) to calibrate and adapt the model: construct a function that outputs calibrated probabilities $p(y|x)$ for each example $x$ in the testing set; \textbf{(2)}  for a given metric of interest, compute an estimate of the improvement in the metric if a particular subset of examples were abstained on by using the calibrated probabilities to specify a categorical distribution over the true labels; we propose efficient algorithms for this step for specific metrics, but in the general case a Monte-Carlo estimate can be used;  \textbf{(3)}  abstain on the subset of examples that gives the largest estimated boost in the performance metric of interest, subject to the constraints of the abstention setting. In our experiments we explore the constraint of abstaining on no more than a fraction $k$ of examples, for the reasons outlined in Sec. \ref{sec:problemformulation}.

In the simplest case where the metric is accuracy, the approach above reduces to abstaining on examples in ascending order of the (calibrated) probability of the most confident class. This is similar to the baseline proposed in \citet{henNgim}, with the minor difference that probabilities are explicitly calibrated and label-shift-adapted.

We noted that step (2) can be obtained via a Monte-Carlo estimate, though this can become expensive if not done carefully. In this work, we propose efficient algorithms for step (2) for three metrics commonly used in settings where label shift is common: the sensitivity at a target level of specificity, the area under the ROC Curve (auROC), and the weighted Cohen's Kappa metric. We discuss them below.

\subsection{Optimizing Sensitivity at a Target Level of Specificity}
\label{sec:optimsensitivityatspecificity}

The sensitivity of a binary classifier is defined as the ratio of correctly-predicted positives to labeled positives, while the specificity of a classifier is defined as one minus the ratio of incorrectely-predicted positives to labeled negatives. A ROC curve is obtained by plotting the sensitivity against one minus the specificity. In disease detection and testing, we typically wish to optimize the sensitivity at a certain specificity level.

\begin{algorithm}[H]
    \small
    \SetAlgoLined
    \SetKwInOut{Input}{input}\SetKwInOut{Output}{output}
    
    \Input{abstention budget $d$, target specificity $s$, sorted calibrated probabilities vector $\mathbf{p}$ of length $N$}\;

    \Output{Vector $\mathbf{o}$ where $o_i$ is estimated sensitivity at specificity $s$ if indices $[i,i+d)$ are excluded}
    \BlankLine
    Initialize $\mathbf{o}$ to a vector of zeros of length $N + 1 - d$ \;
    
    \For{$m \leftarrow 1$ \KwTo $M$}{
        1)  Sample labels vector $\mathbf{y}$ w.r.t. the probabilities $\mathbf{p}$\;
        
        2)  Compute $\mathbf{n}^+_i \coloneqq \sum_{j \ge i} y_{j}$ and $\mathbf{n}^-_i \coloneqq \sum_{j \ge i} 1-y_{j}$.\;
        
        3)  Compute $t^* \coloneqq \min \{ i | 1 - n^-_i/n^-_0 \ge s\}$.\;
        
        4)  Compute $t^\leftarrow_j \coloneqq \min \{i \mid 1 - (n^-_i)/(n^-_0 - j) \ge s\}$ and $t^\rightarrow_j \coloneqq \min \{ i \mid 1 - (n^-_i-j)/(n^-_0 - j) \ge s\}$.\;
        
        5)  Compute $w^-_i \coloneqq n^-_i - n^-_{i+d}$ and $w^+_i \coloneqq n^+_i - n^+_{i+d}$.\;
        
        6) 
        \For{$i \leftarrow 0$ \KwTo $N-d$}{
            Calculate
            $t'_i := \mathbbm{1}\{ t^\rightarrow_{w^-_i} \le i\} t^\rightarrow_{w^-_i} + \mathbbm{1}\{ t^\rightarrow_{w^-_i} > i\} \max(t^\leftarrow_{w^-_i}, i+d)$\;
            
            Update $o_i \leftarrow o_i + (n^+_{t'_i} - \mathbbm{1}\{ t'_i \le i\} w^+_i)/(n^+_0 - w^+_i)$
        }
    }
    Normalize using $o_i \leftarrow o_i/M$.
     \caption{\small Optimizing Sensitivity at Target Specificity}
     \label{alg:sensatspec}
\end{algorithm}

Consider the bounded abstention case of abstaining on at most $d$ examples. Let $N$ denote the total number of examples, and let $\mathbf{p}$ denote the vector of calibrated predicted probabilities sorted such that $p_i \le p_{i+1} \forall i$. Let $[I_0, I_0 + d)$ denote an abstention interval on $p_i$, i.e. all indices $ I_0 \le i < I_0 + d$ are abstained on. How can we apply the framework proposed in Sec. ~\ref{sec:frameworkoverview} to select the optimal interval $[I_0, I_0 + d)$? A naive Monte Carlo algorithm would proceed as follows: for every possible value of $I_0 \in [0, N - d]$, take $M$ Monte-Carlo samples of the label vector $\mathbf{y}$ and calculate the target metric for the set of examples remaining after all indices $i \in [I_0, I_0 + d)$ are excluded. Assuming the metric of interest can be computed in $O(N)$ time (as is the case for sensitivity at a target specificity when the vector of predictions $\mathbf{p}$ is sorted), this algorithm has a runtime of $O((N-d+1)NM) = O(N^2M)$. 

How can we improve on the runtime $O(N^2M)$? For optimizing sensitivity at a target specificity, we devised Algorithm ~\ref{alg:sensatspec}, which has a runtime of $O(NM)$. An (anonymized) video explaining Algorithm ~\ref{alg:sensatspec} is here: \url{https://youtu.be/gpvSMKvrddo}. A runtime analysis for Algorithm ~\ref{alg:sensatspec} is in Sec. ~\ref{sec:runtime_analysis}.

\subsection{Optimizing auROC}
\label{sec:optimauroc}

In Sec. ~\ref{sec:optimsensitivityatspecificity}, we considered the optimization of a specific point on the ROC curve. However, in some applications we may be interested in the overall area under the ROC curve (auROC). This is equal to the probability that a randomly chosen positive will be ranked above a randomly chosen negative \citep{Hanley1982-ll}. While we show there exists an $O(NM)$ Monte Carlo abstention algorithm for the auROC, we obtained strong results with an even more efficient $O(N)$ algorithm. Our main insight was to substitute the calibrated class probabilities for their labels, bypassing the need for Monte Carlo sampling. We discuss both algorithms, as well as their empirical performance, in Sec. ~\ref{sec:appendixoptimauroc}. 

\subsection{Optimizing Weighted Cohen's Kappa}
\label{sec:optimcohenskappa}

Cohen's Kappa quantifies the the agreement between two sets of ratings on categorical classes. Consider the multiclass setting where a predictor $f$ outputs a class $f(x)$ given an example $x$ (in neural networks, this is typically done by taking the $\argmax$ over the predicted class probabilities). Let $S$ denote the set of all examples, $N$ denote the size of $S$, $y(x)$ denote the true class of example $x$, and $W_{i,j}$ denote the penalty for predicting an example from class $i$ as being of class $j$. Let $N^i = \sum_{x \in S} \mathbbm{1}\{ y(x) = i\}$ denote the true number of examples in class $i$, and $F^i \coloneqq \sum_S \mathbbm{1}\{f(x) = i\}$ denote the number of examples predicted by $f$ as as being in class $i$. The weighted Cohen's Kappa metric is defined as $\kappa(S,f) \coloneqq 1 - \frac{\sum_{x \in S} W_{y(x),f(x)}}{\sum_i \sum_j \left( W_{i,j} \frac{N^i}{N} F^j \right)}$. A $\kappa$ of $1$ indicates perfect agreement between the ground truth ratings and $f$, while a $\kappa$ of near $0$ indicates random agreement (i.e. the level of agreement that would be expected if the class proportions produced by $f$ were kept fixed, but the predictions were made at random). A $\kappa$ of less than $0$ indicates that the agreement is worse than what is expected if predictions were made random.

In the binary cases of Sec. ~\ref{sec:optimsensitivityatspecificity} \& ~\ref{sec:optimauroc}, we denoted a contiguous region of abstention by specifying the left index of the interval $I_0$ and the size of the interval $d$. In the multiclass case, it is not as straightforward to specify a contiguous abstention region as the number of possible regions grows exponentially in the number of dimensions. We circumvent this by estimating the \emph{marginal} improvement in the Cohen's Kappa metric when a single example is abstained on, and then abstaining on the subset of examples with the highest total estimated marginal improvement. This approach produced strong empirical results (Tables ~\ref{tab:dr_yesweightrescale_nolabelshift}, ~\ref{tab:dr_yesweightrescale_yeslabelshift}, ~\ref{tab:dr_noweightrescale_nolabelshift} \& ~\ref{tab:dr_noweightrescale_yeslabelshift}). We developed a deterministic $O(NC)$ algorithm for estimating the marginal improvements for all examples, where $C$ is the number of classes. The algorithm, as well as empirical analysis of convergence, are detailed in Sec. ~\ref{sec:appendixoptimcohenskappa}.

\subsection{Additional Baseline: JS Divergence from Prior Class Probabilities}
\label{sec:js-div}

We additionally present a novel baseline that to our knowledge has not been discussed in the literature: abstaining on those examples for which the Jensen-Shannon (JS) divergence of the predicted probabilities is most similar to the prior class probabilities. In particular, the lower the JS divergence between each example's calibrated predicted class probabilities and the prior class probabilities, the closer the prediction to the prior class proportions, and therefore the less information we have about the prediction. Although this strategy does not give consideration to the specific performance metric of interest, it is computationally tractable and adaptable to label shift in the testing set.

\section{Experiments}
\label{sec:experiments}
We studied the behaviour of the methods described in Sec. ~\ref{sec:preexistingbaselines} and our proposed methods in Sec. ~\ref{sec:frameworkoverview} \& ~\ref{sec:js-div} on test set distributions both with and without simulated label shift. Calibration and label-shift adaptation, where applicable, were performed using the equivalent of a heldout validation set. The abstention methods were never presented with the test-set labels. For a given model, dataset, and abstention method, the performance of the abstention method was quantified by the improvement in the metric of interest when different percentages of the dataset were abstained on (denoted as ``@$x$\% Abst.'' in the column headings). To assess whether one abstention method was significantly better than another, we applied a one-sided Wilcoxon signed rank test at a significance threshold of $0.05$ to a distribution of performance values generated using models trained with different random seeds as well as different resamplings of the validation and testing sets  (specifics are detailed in table captions). We chose the Wilcoxon signed rank test because it is designed to handle paired samples without making any other assumptions about the distribution (a paired sample test was necessary because performance is dependant upon the specific model and data used).

Since calibration and label-shift adaptation were not specified as a preprocessing step in the baselines of Sec. ~\ref{sec:preexistingbaselines}, we ran all baselines both with and without calibration in the experiments without label shift (Tab. \ref{tab:heme_task1}, \ref{tab:heme_task0}, \ref{tab:dr_yesweightrescale_nolabelshift} \& \ref{tab:dr_noweightrescale_nolabelshift}). We found that having calibrated probabilities improves the baselines. For this reason, in the experiments that did involve label shift (Tab. \ref{tab:imdb_and_catdog}, \ref{tab:dr_yesweightrescale_yeslabelshift} \& \ref{tab:dr_noweightrescale_yeslabelshift}), we ran all methods using calibrated probabilities.

\textbf{Code (anonymized):} code for the algorithms is at \url{https://github.com/blindauth/abstention}, and notebooks for the experiments are at \url{https://github.com/blindauth/abstention_experiments}.

\begin{table*}[!ht]
\tiny
\begin{center}
\begin{tabular}{ | c c | c c c | c c c | }
\hline
\multicolumn{2}{| c |}{ } & \multicolumn{3}{ c |}{IMDB Sensitivity @ 99\% Specificity} & \multicolumn{3}{c | }{Cat v Dog Sensitivity @ 99\% Specificity}\\
\multicolumn{1}{| c}{Method} & \multicolumn{1}{c |}{Adapted?} & Base & @30\% Abst. & @15\% Abst. & Base & @30\% Abst. & @15\% Abst.\\ \hline
Est $\Delta$Metric (Ours) & Y & 0.3571$\pm$0.004 & \textbf{0.7067$\pm$0.0067} & \textbf{0.5243$\pm$0.0059} & 0.3318$\pm$0.0061 & \textbf{0.6135$\pm$0.0114} & \textbf{0.4598$\pm$0.0082}\\ \hline
Est $\Delta$Metric (Ours) & N & 0.3571$\pm$0.004 & 0.6227$\pm$0.0082 & 0.4959$\pm$0.0057 & 0.3318$\pm$0.0061 & 0.5616$\pm$0.0102 & 0.4438$\pm$0.0084\\ \hline
JS Div. (Ours) & Y & 0.3571$\pm$0.004 & 0.4175$\pm$0.0046 & 0.3884$\pm$0.0048 & 0.3318$\pm$0.0061 & 0.404$\pm$0.0079 & 0.3655$\pm$0.0064\\ \hline
JS Div. (Ours) & N & 0.3571$\pm$0.004 & 0.3869$\pm$0.0045 & 0.3722$\pm$0.0043 & 0.3318$\pm$0.0061 & 0.3687$\pm$0.0073 & 0.3466$\pm$0.0065\\ \hline
\hline
Dist. from 0.5 & Y & 0.3571$\pm$0.004 & 0.5552$\pm$0.0071 & 0.4317$\pm$0.0052 & 0.3318$\pm$0.0061 & 0.5226$\pm$0.01 & 0.4122$\pm$0.0078\\ \hline
Dist. from 0.5 & N & 0.3571$\pm$0.004 & 0.4428$\pm$0.0052 & 0.3909$\pm$0.0048 & 0.3318$\pm$0.0061 & 0.4134$\pm$0.0079 & 0.3665$\pm$0.0064\\ \hline
Fumera et al. & Y & 0.3571$\pm$0.004 & 0.5869$\pm$0.008 & 0.493$\pm$0.0059 & 0.3318$\pm$0.0061 & 0.529$\pm$0.0099 & 0.4453$\pm$0.0084\\ \hline
Fumera et al. & N & 0.3571$\pm$0.004 & 0.6976$\pm$0.0079 & 0.441$\pm$0.0054 & 0.3318$\pm$0.0061 & \textbf{0.6154$\pm$0.0103} & 0.4017$\pm$0.0078\\ \hline
Test-time Dropout & Y & 0.3571$\pm$0.004 & 0.5751$\pm$0.008 & 0.4407$\pm$0.0051 & 0.3318$\pm$0.0061 & 0.4522$\pm$0.0137 & 0.3748$\pm$0.0083\\ \hline
Test-time Dropout & N & 0.3571$\pm$0.004 & 0.4625$\pm$0.0055 & 0.3994$\pm$0.0042 & 0.3318$\pm$0.0061 & 0.3747$\pm$0.0099 & 0.3492$\pm$0.0074\\ \hline
\end{tabular}
\end{center}
\caption{\small \textbf{Sensitivity at 99\% specificity for IMDB sentiment classification and Cat v Dog image classification under label shift}.  The training and validation sets had the original 1:1 ratio of positives:negatives, while the test set had a positives:negatives ratio of 1:2. Values represent the mean and standard error across samples generated using a combination of different model initializations and different bootstrapped versions of the heldout sets. ``Base'' indicates performance without abstention. ``Adapted?'' indicates whether predicted probabilities were adapted to account for label shift using the EM approach of \citet{Saerens2002-jh}. All probabilities were calibrated using Platt scaling \citep{platt} on the validation set. Within a column, bold values are significantly better than non-bold values according to a Wilcoxon signed rank test with p < 0.05. Bold values are not significantly different from other bold values in the same column. Note that label shift adaptation can \emph{worsen} the approach of Fumera et al., likely because that approach tailors its abstention boundaries to the validation set. See Sec. ~\ref{sec:imdbwithlabelshift} and Sec. ~\ref{sec:catvdogwithlabelshift} for more details on the data.}
\label{tab:imdb_and_catdog}
\end{table*}

\begin{table*}[!ht]
\tiny
\begin{center}
\begin{tabular}{ | c c | c c c | c c c | }
\hline
\multicolumn{2}{| c |}{ } & \multicolumn{3}{ c |}{
Sensitivity @ 80\% Specificity (Leukemia Stem Cells)} & \multicolumn{3}{c | }{Area under ROC (Leukemia Stem Cells)}\\
\multicolumn{1}{| c}{Method} & \multicolumn{1}{c |}{Calib?} & Base & @30\% Abst. & @15\% Abst. & Base & @30\% Abst. & @15\% Abst.\\ \hline
Est $\Delta$Metric (Ours) & Y & 0.6507$\pm$0.0029 & \textbf{0.8001$\pm$0.0035} & \textbf{0.7279$\pm$0.0033} & 0.8107$\pm$0.0014 & \textbf{0.8586$\pm$0.0015} & \textbf{0.8344$\pm$0.0014}\\ \hline
JS Div. (Ours) & Y & 0.6507$\pm$0.0029 & 0.7907$\pm$0.0041 & 0.7177$\pm$0.0037 & 0.8107$\pm$0.0014 & 0.8578$\pm$0.0015 & 0.8342$\pm$0.0014\\ \hline
\hline
Dist. from 0.5 & Y & 0.6507$\pm$0.0029 & 0.7108$\pm$0.0027 & 0.6825$\pm$0.0037 & 0.8107$\pm$0.0014 & 0.8415$\pm$0.0014 & 0.8259$\pm$0.0016\\ \hline
Dist. from 0.5 & N & 0.6507$\pm$0.0029 & 0.7374$\pm$0.0084 & 0.6934$\pm$0.0057 & 0.8107$\pm$0.0014 & 0.8474$\pm$0.0033 & 0.8289$\pm$0.0022\\ \hline
Fumera et al. & Y & 0.6507$\pm$0.0029 & 0.7923$\pm$0.0038 & 0.7111$\pm$0.0033 & 0.8107$\pm$0.0014 & 0.8581$\pm$0.0016 & 0.8327$\pm$0.0015\\ \hline
Fumera et al. & N & 0.6507$\pm$0.0029 & 0.7882$\pm$0.0059 & 0.7132$\pm$0.0054 & 0.8107$\pm$0.0014 & 0.8573$\pm$0.0018 & 0.8321$\pm$0.0019\\ \hline
Test-time Dropout & Y & 0.6507$\pm$0.0029 & 0.6768$\pm$0.0096 & 0.6688$\pm$0.0052 & 0.8107$\pm$0.0014 & 0.8216$\pm$0.0039 & 0.8177$\pm$0.0021\\ \hline
Test-time Dropout & N & 0.6507$\pm$0.0029 & 0.7006$\pm$0.008 & 0.6787$\pm$0.0051 & 0.8107$\pm$0.0014 & 0.8305$\pm$0.0039 & 0.8215$\pm$0.0026\\ \hline
\end{tabular}
\end{center}
\caption{\small \textbf{Sensitivity at 80\% specificity and auROC for identifying regions active in Leukemia Stem Cells}. Values represent the mean and standard error across samples generated using a combination of different model initializations and different random splits of the heldout set. ``Base'' indicates performance without abstention. ``Calib?'' indicates whether predicted probabilities were calibrated using Platt scaling \citep{platt}. Within a column, bold values are significantly better than non-bold values according to a Wilcoxon signed rank test (p < 0.05). Ratio of positives:negatives was roughly 1:2. See Sec. ~\ref{sec:genomicsdataset} for more details on the data. Analogous results for genomic regions active in preleukemic hematopoetic stem cells (pHSCs) are in Table ~\ref{tab:heme_task0}}
\label{tab:heme_task1}
\end{table*}

\begin{table*}[!ht]
\tiny
\begin{center}
\begin{tabular}{ | c c | c c c c c c | }
\hline
\multicolumn{2}{| c |}{ } & \multicolumn{6}{ c |}{Diabetic Retinopathy Cohen's Weighted Kappa (no label shift)}\\
\multicolumn{1}{| c}{Method} & \multicolumn{1}{c |}{Calib?} & Base & @30\% Abst. & @25\% Abst. & @20\% Abst. & @15\% Abst. & @10\% Abst. \\ \hline
Est $\Delta$Metric (Ours) & Y & 0.8104$\pm$0.0019 & \textbf{0.8918$\pm$0.0013} & \textbf{0.8875$\pm$0.0014} & \textbf{0.88$\pm$0.0016} & \textbf{0.8714$\pm$0.0016} & \textbf{0.8597$\pm$0.0017}\\ \hline
JS Div. (Ours) & Y & 0.8104$\pm$0.0019 & 0.8529$\pm$0.0015 & 0.8497$\pm$0.0015 & 0.8463$\pm$0.0015 & 0.8419$\pm$0.0015 & 0.8357$\pm$0.0015\\ \hline
\hline
Entropy & Y & 0.8104$\pm$0.0019 & 0.7667$\pm$0.0064 & 0.8039$\pm$0.0047 & 0.8318$\pm$0.0032 & 0.8433$\pm$0.0025 & 0.8417$\pm$0.002\\ \hline
Entropy & N & 0.8104$\pm$0.0019 & 0.7517$\pm$0.0067 & 0.7883$\pm$0.0049 & 0.814$\pm$0.0035 & 0.831$\pm$0.0028 & 0.8378$\pm$0.0022\\ \hline
Max Class Prob. & Y & 0.8104$\pm$0.0019 & 0.7522$\pm$0.0067 & 0.7879$\pm$0.0049 & 0.8107$\pm$0.0035 & 0.8236$\pm$0.0029 & 0.8279$\pm$0.0023\\ \hline
Max Class Prob. & N & 0.8104$\pm$0.0019 & 0.7527$\pm$0.0066 & 0.7856$\pm$0.0048 & 0.8046$\pm$0.0036 & 0.8189$\pm$0.0029 & 0.8241$\pm$0.0024\\ \hline
Test-time Dropout & Y & 0.8104$\pm$0.0019 & 0.6535$\pm$0.0099 & 0.706$\pm$0.0071 & 0.7428$\pm$0.0051 & 0.7694$\pm$0.0037 & 0.7849$\pm$0.003\\ \hline
Test-time Dropout & N & 0.8104$\pm$0.0019 & 0.6775$\pm$0.0101 & 0.73$\pm$0.0067 & 0.7575$\pm$0.0049 & 0.777$\pm$0.0038 & 0.7872$\pm$0.003\\ \hline
\end{tabular}
\end{center}
\caption{\small \textbf{Cohen's Weighted Kappa for Diabetic Retinopathy Detection (no label shift)}. Values represent the mean and standard error across samples generated using different random splits of the heldout sets. ``Base'' indicates performance without abstention. ``Calib?'' indicates whether predicted probabilities were calibrated using Temperature scaling \citep{calibration}. Within a column, bold values are significantly better than non-bold values according to a Wilcoxon signed rank test (p < 0.05). See Sec. ~\ref{sec:diabeticretionopathydata} for more details on the data. See Table ~\ref{tab:dr_yesweightrescale_yeslabelshift} for corresponding experiments under simulated label shift, and Tables ~\ref{tab:dr_noweightrescale_nolabelshift} \& ~\ref{tab:dr_noweightrescale_yeslabelshift} for experiments exploring the effect of using the expected value over several dropout runs rather than deterministic dropout at test-time.}
\label{tab:dr_yesweightrescale_nolabelshift}
\end{table*}

\begin{table*}[!ht]
\tiny
\begin{center}
\begin{tabular}{ | c c | c c c c c c | }
\hline
\multicolumn{2}{| c |}{ } & \multicolumn{6}{ c |}{Diabetic Retinopathy Cohen's Weighted Kappa (with label shift)}\\
\multicolumn{1}{| c}{Method} & \multicolumn{1}{c |}{Adapted?} & Base & @30\% Abst. & @25\% Abst. & @20\% Abst. & @15\% Abst. & @10\% Abst. \\ \hline
Est $\Delta$Metric (Ours) & Y & 0.7817$\pm$0.0045 & \textbf{0.8559$\pm$0.0027} & \textbf{0.8499$\pm$0.0028} & \textbf{0.8423$\pm$0.003} & \textbf{0.8319$\pm$0.0034} & \textbf{0.8203$\pm$0.0037}\\ \hline
Est $\Delta$Metric (Ours) & N & 0.732$\pm$0.0039 & 0.8228$\pm$0.0036 & 0.8155$\pm$0.0035 & 0.8061$\pm$0.0037 & 0.7968$\pm$0.0036 & 0.7837$\pm$0.0038\\ \hline
JS Div. (Ours) & Y & 0.7817$\pm$0.0045 & 0.8396$\pm$0.0028 & 0.84$\pm$0.0028 & 0.838$\pm$0.0028 & \textbf{0.8316$\pm$0.0032} & \textbf{0.8203$\pm$0.0036}\\ \hline
JS Div. (Ours) & N & 0.732$\pm$0.0039 & 0.8457$\pm$0.003 & 0.8397$\pm$0.0031 & 0.8288$\pm$0.0032 & 0.8154$\pm$0.0033 & 0.7972$\pm$0.0035\\ \hline
\hline
Entropy & Y & 0.7817$\pm$0.0045 & 0.8426$\pm$0.0041 & 0.8346$\pm$0.0043 & 0.8274$\pm$0.0045 & 0.8191$\pm$0.0046 & 0.8086$\pm$0.0047\\ \hline
Entropy & N & 0.732$\pm$0.0039 & 0.7837$\pm$0.0053 & 0.7795$\pm$0.0049 & 0.7756$\pm$0.0046 & 0.7697$\pm$0.0044 & 0.7602$\pm$0.004\\ \hline
Max Class Prob. & Y & 0.7817$\pm$0.0045 & 0.8339$\pm$0.005 & 0.8261$\pm$0.0049 & 0.819$\pm$0.0048 & 0.8117$\pm$0.0048 & 0.8029$\pm$0.0048\\ \hline
Max Class Prob. & N & 0.732$\pm$0.0039 & 0.7788$\pm$0.0054 & 0.7727$\pm$0.0049 & 0.7678$\pm$0.0045 & 0.7625$\pm$0.0044 & 0.7533$\pm$0.0042\\ \hline
Test-time Dropout & Y & 0.7817$\pm$0.0045 & 0.7912$\pm$0.0047 & 0.7843$\pm$0.0045 & 0.7776$\pm$0.0043 & 0.7743$\pm$0.0043 & 0.7718$\pm$0.0044\\ \hline
Test-time Dropout & N & 0.732$\pm$0.0039 & 0.7532$\pm$0.0054 & 0.7456$\pm$0.005 & 0.7399$\pm$0.0047 & 0.7334$\pm$0.0043 & 0.7236$\pm$0.0045\\ \hline
\end{tabular}
\end{center}
\caption{\small \textbf{Cohen's Weighted Kappa for Diabetic Retinopathy Detection, with label shift}. Setup is similar to Table ~\ref{tab:dr_yesweightrescale_nolabelshift}. ``Adapted?'' indicates whether predicted probabilities were adapted to account for label shift using the EM approach of \citet{Saerens2002-jh}. All probabilities were calibrated using bias-corrected Temperature Scaling \citep{biascorrectedtempscaling}.}
\label{tab:dr_yesweightrescale_yeslabelshift}
\end{table*}

\subsection{Binary Simulated, Sentiment, and Image Classification}
\label{sec:simulatedbinarydataset}
\label{sec:imdbwithlabelshift}
\label{sec:catvdogwithlabelshift}

\textbf{Simulated Binary Classification}:  We simulated the scalar predictions $x_i$ of a classifier as follows: the true class $y_i$ of example $i$ was determined by drawing from a Bernoulli distribution where the probability of the positive class is $q$. If the example was a positive, $x_i$ was sampled from a normal distribution with a mean of $\mu^+$ and a standard deviation of $\sigma^+$. Otherwise, $x_i$ was sampled from a normal distribution with a mean of $\mu^-$ and a standard deviation of $\sigma^-$. Once $x_i$ was sampled, we analytically calculated the calibrated posterior probability that the $x_i$ originated from the positive class (given only the observed value of $x_i$)  according to the parameters of the data generating distribution. A total of $n$ examples were sampled in this way. For Figure ~\ref{fig:importanceofmetricspecific}, we used $q = 0.1$, $\mu^- = -1$, $\mu^+ = 2$, $\sigma^- = 2$, $\sigma^+ = 1$ and $n=10000$. Abstention was then performed as described in the caption of Figure ~\ref{fig:importanceofmetricspecific}.

\textbf{Sentiment Classification}:  We used the large movie review (IMDB) sentiment classification dataset \citep{imdb} and the corresponding convolutional neural network model that ships with Keras \citep{keras}. The original test set contained 20K examples, and the original validation set contained 10K examples. Label shift was simulated by sampling 10K examples from the test set at an imbalance ratio of 1:2 in favor of negatives. Ten models were trained (each with different random seeds), and for each model, three different bootstrapped samples of the validation set and label-shifted test set were generated. This resulted in a total of 30 experiments on which to perform statistical comparisons. Results are shown in Table ~\ref{tab:imdb_and_catdog}.

\textbf{Image Classification}:  We trained a CNN to distinguish images of cats and dogs using the ASIRRA dataset \citep{asirra}. Images were resized to have dimensions 64 $\times$ 64 $\times$ 3. The original test and validation sets contained 5K examples each. Label shift was simulated by sampling 5K examples from the test set at an imbalance ratio of 1:2 in favor of cats (cats were labeled as the negatives). The setup was otherwise similar to what was used for the sentiment classification task. Results are shown in Table ~\ref{tab:imdb_and_catdog}.

\subsection{Identifying Active Regions in Leukemic and Preleukemic Stem Cells}
\label{sec:genomicsdataset}

Predicting regulatory activity of non-coding DNA sequences is a complex task. Recently, CNN models have been proposed to predict chromatin accessibility (a biochemical marker of regulatory DNA) in different cell types from DNA sequences. While the label shift assumption does not hold for this problem, the class priors are critically imbalanced and even the best computational models achieve only fair performance. We trained a multi-task Basset model \citep{basset} to map 4-channel (A, C, G, T) one-hot encoded DNA sequences to binary chromatin accessibility outputs across 16 hematopoietic cell types (16 binary classification tasks) \citep{hema-dataset}. The dataset contained 837,977 sequences underlying in-vivo chromatin accessible sites across all 16 cell types. We evaluated the methods on two tasks (cell types): preleukemic hematopoetic stem cells (pHSCs) and leukemia stem cells (LSCs). The ratio of negatives:positives for both tasks was roughly 2:1. Ten models were trained with different random seeds. We sampled three different random splits into the validation and testing set per model, giving a total of $10\times3 = 30$ experiments on which to perform statistical comparisons. Results for LSCs and pHSCs are in Tables ~\ref{tab:heme_task1} and ~\ref{tab:heme_task0} respectively.

\subsection{Diabetic Retinopathy Detection}
\label{sec:diabeticretionopathydata}

This dataset consists of high-resolution retinal images that are graded on an integer scale ranging from 0-4, with 0 indicating ``No Diabetic Retinopathy'' and 4 indicating ``Proliferative Diabetic Retinopathy'' (most severe). Performance is evaluated using Cohen's weighted Kappa metric with $W_{i,j} = (i-j)^2$. i.e., incorrect predictions are penalized more strongly if the difference between the true and the predicted ratings is larger.

We used the publicly available pre-trained model from \citet{JeffreyD27:online}. We performed our assessment by evenly dividing the 3514 held-out images used for evaluation by \citet{JeffreyD27:online} into a ``calibration set'' (to calibrate the predictor) and an ``evaluation set'' (to compute Cohen's Weighted Kappa after abstention). We were careful to keep images of the left and right eyes that originated from the same patient in the same set, and kept the class proportions roughly consistent between the two sets. For statistical comparisons using the Wilcoxon test, we sampled 30 different splits into calibration sets and evaluation sets. Images within a set were further augmented by a factor of 8 with horizontal flipping and rotation (the factor of 8 comes from having two possible options for flip vs. no flip, multiplied by four possible rotation amounts - \ang{0}, \ang{90}, \ang{180} and \ang{270}). Results for optimizing Cohen's Weighted Kappa in the absence of label shift are shown in Tables ~\ref{tab:dr_yesweightrescale_nolabelshift} \& ~\ref{tab:dr_noweightrescale_nolabelshift}. 

For the experiments involving label shift on the Diabetic Retinopathy (DR) dataset, rather than augmenting every image in the ``evaluation set'' by a factor of 8, we augmented images by different amounts depending on their class. Images of grade 0 (``No DR'') were not augmented, images of grade 1 (``Mild DR'') were augmented by a factor of 2, images of grade 2 (``Moderate DR'') were augmented by a factor of 5, and images of grades 3 \& 4 (``Severe DR'' and ``Proliferative DR'') were augmented by a factor of 8. Results for optimizing Cohen's Weighted Kappa in the presence of label shift are shown in Tables ~\ref{tab:dr_yesweightrescale_yeslabelshift} \& ~\ref{tab:dr_noweightrescale_yeslabelshift}.

\section{Discussion}
\label{sec:discussion}

In this work, we used calibrated and label-shift-adapted probabilities from a classifier to specify a categorical distribution over the labels in the test set. Given any metric of interest, we showed how we could use this distribution to estimate the improvement in the metric if a particular set of examples were abstained on. Under this framework, we devised efficient algorithms to optimize metrics typically deployed in practical settings, such as: sensitivity at target specificity, auROC, and Cohen's weighted Kappa. In our experiments, we demonstrated the advantage of this approach using simulated label shift (Tables ~\ref{tab:imdb_and_catdog},  ~\ref{tab:dr_yesweightrescale_yeslabelshift} \& ~\ref{tab:dr_noweightrescale_yeslabelshift}). We observed that the proposed approach was uniformly better than other methods, both in the presence and absence of label shift, and across different abstention constraints.

Furthermore, in the absence of a specific metric of interest, we showed that abstention based on JS divergence from the prior class proportions can be a surprisingly effective tool. In several experiments, simply abstaining on cases that had calibrated predicted probabilities closest to the prior class probabilities as per JS divergence worked reasonably well without optimizing for the specific metric. This finding suggests that abstaining based on JS divergence can an appropriate first strategy for practitioners and a formidable baseline for researchers to compare against.

In some scenarios, the abstention cost may not be monotonically increasing in the total number of abstentions because the cost may depend on the true class of the abstained example. Our framework can be extended to this cost-based case by again using the calibrated probabilities as a proxy for the true labels, and then estimating the cost of abstaining on a given subset of examples analogously to how the improvement in a metric is estimated. We leave the detailed explorations of this direction for future work.

More broadly, we showed that abstention approaches that rely on predicted probabilities, whether off-the-shelf or tailored to a specific metric, should employ calibration and label shift adaptation; with our analysis, we show that label shift adaptation tends to improve not just our methods but also several of the other baselines we considered. There is one notable exception to this: the method of \citet{fumeramultiplethresholds}, which involved a brute-force search on the validation set to identify abstention thresholds. However, there is no guarantee that validation set thresholds are optimal for a test set with different prior probabilities.

Systems discussed here, such as medical diagnostic systems, can cause injury if they malfunction \citep{bowen1993safety}. Moreover, they are subject to rigorous inspection and testing, e.g. FDA certification. Consequently, contributing principled and intuitive tools for practitioners to perform abstention has the potential to provide a strong positive impact. The improved computational efficiency could result in wider deployment and faster flagging of difficult cases. Still, more work needs to be done to standardize specific tasks by regulators and technology-stakeholders. Nevertheless, we believe that our method will enhance the robustness of diagnostic systems, and we hope this work will inspire practitioners to leverage more appropriate abstention algorithms in applications where metrics other than accuracy are used or where the label shift assumption holds.

\FloatBarrier

\newpage

\bibliography{main}

\begin{thebibliography}{43}
\providecommand{\natexlab}[1]{#1}
\providecommand{\url}[1]{\texttt{#1}}
\expandafter\ifx\csname urlstyle\endcsname\relax
  \providecommand{\doi}[1]{doi: #1}\else
  \providecommand{\doi}{doi: \begingroup \urlstyle{rm}\Url}\fi

\bibitem[Bay()]{Bayesian0:online}
Bayesian deep learning - quantopian blog.
\newblock \url{https://blog.quantopian.com/bayesian-deep-learning/}.
\newblock (Accessed on 05/17/2018).

\bibitem[Alexandari et~al.(2020)Alexandari, Kundaje, and
  Shrikumar]{biascorrectedtempscaling}
Amr Alexandari, Anshul Kundaje, and Avanti Shrikumar.
\newblock Maximum likelihood with bias-corrected calibration is hard-to-beat at
  label shift adaptation.
\newblock In \emph{International Conference on Machine Learning}, pages
  222--232. PMLR, 2020.

\bibitem[Awate et~al.(2019)Awate, Garg, and Jena]{awate2019estimating}
Suyash~P Awate, Saurabh Garg, and Rohit Jena.
\newblock Estimating uncertainty in mrf-based image segmentation: A
  perfect-mcmc approach.
\newblock \emph{Medical image analysis}, 55:\penalty0 181--196, 2019.

\bibitem[Bartlett and Wegkamp(2008)]{Bartlett2008-qf}
Peter~L Bartlett and Marten~H Wegkamp.
\newblock Classification with a reject option using a hinge loss.
\newblock \emph{J. Mach. Learn. Res.}, 9\penalty0 (Aug):\penalty0 1823--1840,
  2008.

\bibitem[Bowen and Stavridou(1993)]{bowen1993safety}
Jonathan Bowen and Victoria Stavridou.
\newblock Safety-critical systems, formal methods and standards.
\newblock \emph{Software engineering journal}, 8\penalty0 (4):\penalty0
  189--209, 1993.

\bibitem[Chollet(2018)]{keras}
F.~Chollet.
\newblock Keras.
\newblock 2018.

\bibitem[Corces et~al.(2016)Corces, Buenrostro, Wu, Greenside, Chan, Koenig,
  Snyder, Pritchard, Kundaje, Greenleaf, Majeti, and Chang]{hema-dataset}
M.~R. Corces, J.~D. Buenrostro, B.~Wu, P.~G. Greenside, S.~M. Chan, J.~L.
  Koenig, M.~P. Snyder, J.~K. Pritchard, A.~Kundaje, W.~J. Greenleaf,
  R.~Majeti, and H.~Y. Chang.
\newblock Lineage-specific and single-cell chromatin accessibility charts human
  hematopoiesis and leukemia evolution.
\newblock \emph{Nature Genetics}, 48\penalty0 (10), 2016.

\bibitem[Cordella et~al.(1995)Cordella, De~Stefano, Tortorella, and
  Vento]{Cordella1995-qy}
L~P Cordella, C~De~Stefano, F~Tortorella, and M~Vento.
\newblock A method for improving classification reliability of multilayer
  perceptrons.
\newblock \emph{IEEE Trans. Neural Netw.}, 6\penalty0 (5):\penalty0 1140--1147,
  1995.

\bibitem[Cortes et~al.(2016)Cortes, DeSalvo, and Mohri]{Cortes2016-cm}
Corinna Cortes, Giulia DeSalvo, and Mehryar Mohri.
\newblock Boosting with abstention.
\newblock In D~D Lee, M~Sugiyama, U~V Luxburg, I~Guyon, and R~Garnett, editors,
  \emph{Advances in Neural Information Processing Systems 29}, pages
  1660--1668. Curran Associates, Inc., 2016.

\bibitem[De~Fauw(2015)]{JeffreyD27:online}
Jeffrey De~Fauw.
\newblock Jeffreydf/kaggle\_diabetic\_retinopathy: Fifth place solution of the
  kaggle diabetic retinopathy competition.
\newblock \url{https://github.com/JeffreyDF/kaggle_diabetic_retinopathy}, Oct
  2015.
\newblock (Accessed on 01/22/2019).

\bibitem[De~Stefano et~al.(2000)De~Stefano, Sansone, and
  Vento]{De_Stefano2000-ub}
Claudio De~Stefano, Carlo Sansone, and Mario Vento.
\newblock To reject or not to reject: that is the question-an answer in case of
  neural classifiers.
\newblock \emph{IEEE Trans. Syst. Man Cybern. C Appl. Rev.}, 30\penalty0
  (1):\penalty0 84--94, 2000.

\bibitem[D{\"u}rr et~al.(2018)D{\"u}rr, Murina, Siegismund, Tolkachev,
  Steigele, and Sick]{Durr2018-yt}
Oliver D{\"u}rr, Elvis Murina, Daniel Siegismund, Vasily Tolkachev, Stephan
  Steigele, and Beate Sick.
\newblock Know when you don't know: A robust deep learning approach in the
  presence of unknown phenotypes.
\newblock \emph{Assay Drug Dev. Technol.}, 16\penalty0 (6):\penalty0 343--349,
  2018.

\bibitem[El-Yaniv and Wiener(2010)]{El-Yaniv2010-xs}
Ran El-Yaniv and Yair Wiener.
\newblock On the foundations of noise-free selective classification.
\newblock \emph{J. Mach. Learn. Res.}, 11\penalty0 (May):\penalty0 1605--1641,
  2010.

\bibitem[Elson et~al.(2007)Elson, Douceur, Howell, and Saul]{asirra}
J.~Elson, J.~R. Douceur, J.~Howell, and J.~Saul.
\newblock Asirra: A captcha that exploits interest-aligned manual image
  categorization.
\newblock \emph{Proceedings of 14th ACM Conference on Computer and
  Communications Security (CCS)}, 2007.

\bibitem[Fumera and Roli(2002)]{Fumera2002-gh}
Giorgio Fumera and Fabio Roli.
\newblock Support vector machines with embedded reject option.
\newblock In \emph{Pattern Recognition with Support Vector Machines}, pages
  68--82. Springer Berlin Heidelberg, 2002.

\bibitem[Fumera et~al.(2000)Fumera, Roli, and
  Giacinto]{fumeramultiplethresholds}
Giorgio Fumera, Fabio Roli, and Giorgio Giacinto.
\newblock Reject option with multiple thresholds.
\newblock \emph{Pattern Recognit.}, 33\penalty0 (12):\penalty0 2099--2101,
  2000.

\bibitem[Gal and Ghahramani(2016)]{testtime}
Y.~Gal and Z.~Ghahramani.
\newblock Dropout as a bayesian approximation: Representing model uncertainty
  in deep learning.
\newblock \emph{International Conference on Machine Learning (ICML)}, 2016.

\bibitem[Gal and Ghahramani(2015)]{Gal2015-sn}
Yarin Gal and Zoubin Ghahramani.
\newblock Bayesian convolutional neural networks with bernoulli approximate
  variational inference.
\newblock June 2015.

\bibitem[Garg et~al.(2020)Garg, Wu, Balakrishnan, and Lipton]{garglabelshift}
Saurabh Garg, Yifan Wu, Sivaraman Balakrishnan, and Zachary Lipton.
\newblock A unified view of label shift estimation.
\newblock In H.~Larochelle, M.~Ranzato, R.~Hadsell, M.~F. Balcan, and H.~Lin,
  editors, \emph{Advances in Neural Information Processing Systems}, volume~33,
  pages 3290--3300. Curran Associates, Inc., 2020.
\newblock URL
  \url{https://proceedings.neurips.cc/paper/2020/file/219e052492f4008818b8adb6366c7ed6-Paper.pdf}.

\bibitem[{Geifman} and {El-Yaniv}(2017)]{selectiveclassification}
Y.~{Geifman} and R.~{El-Yaniv}.
\newblock {Selective Classification for Deep Neural Networks}.
\newblock \emph{ArXiv e-prints}, 2017.

\bibitem[Guo et~al.(2017)Guo, Pleiss, Sun, and Weinberger]{calibration}
C.~Guo, G.~Pleiss, Y.~Sun, and K.~Q. Weinberger.
\newblock On calibration of modern neural networks.
\newblock \emph{International Conference on Machine Learning (ICML)}, 2017.

\bibitem[Hanley and McNeil(1982)]{Hanley1982-ll}
J~A Hanley and B~J McNeil.
\newblock The meaning and use of the area under a receiver operating
  characteristic ({ROC}) curve.
\newblock \emph{Radiology}, 143\penalty0 (1):\penalty0 29--36, April 1982.

\bibitem[Hellman(1970)]{Hellman1970-zo}
Martin~E Hellman.
\newblock The nearest neighbor classification rule with a reject option.
\newblock \emph{IEEE Transactions on Systems Science and Cybernetics},
  6\penalty0 (3):\penalty0 179--185, 1970.

\bibitem[Hendrycks and Gimpel(2017)]{henNgim}
D.~Hendrycks and K.~Gimpel.
\newblock A baseline for detecting misclassified and out-of-distribution
  examples in neural networks.
\newblock \emph{International Conference on Learning Representations (ICLR)},
  2017.

\bibitem[Jiang et~al.(2012)Jiang, Osl, Kim, and Ohno-Machado]{medical}
X.~Jiang, M.~Osl, J.~Kim, and L.~Ohno-Machado.
\newblock Calibrating predictive model estimates to support personalized
  medicine.
\newblock \emph{Journal of the American Medical Informatics Association},
  19\penalty0 (2):\penalty0 263--274, 2012.

\bibitem[Jones et~al.(2020)Jones, Sagawa, Koh, Kumar, and
  Liang]{jones2020selective}
Erik Jones, Shiori Sagawa, Pang~Wei Koh, Ananya Kumar, and Percy Liang.
\newblock Selective classification can magnify disparities across groups.
\newblock \emph{arXiv preprint arXiv:2010.14134}, 2020.

\bibitem[Kahneman and Tversky(1973)]{kahneman1}
Daniel Kahneman and Amos Tversky.
\newblock On the psychology of prediction.
\newblock \emph{Psychological review}, 80\penalty0 (4):\penalty0 237, 1973.

\bibitem[Kahneman and Tversky(1996)]{kahneman2}
Daniel Kahneman and Amos Tversky.
\newblock On the reality of cognitive illusions.
\newblock 1996.

\bibitem[Kelley et~al.(2016)Kelley, Snoek, and Rinn]{basset}
D.~R. Kelley, J.~Snoek, and J.~Rinn.
\newblock Basset: Learning the regulatory code of the accessible genome with
  deep convolutional neural networks.
\newblock \emph{Genome Research}, 2016.

\bibitem[Krueger et~al.(2018)Krueger, Huang, Islam, Turner, Lacoste, and
  Courville]{bayesianhypernets}
David Krueger, Chin-Wei Huang, Riashat Islam, Ryan Turner, Alexandre Lacoste,
  and Aaron Courville.
\newblock {Bayesian Hypernetworks}.
\newblock \emph{ArXiv e-prints}, 2018.

\bibitem[Leibig et~al.(2017)Leibig, Allken, Ayhan, Berens, and
  Wahl]{Leibig2017-ie}
Christian Leibig, Vaneeda Allken, Murat~Se{\c c}kin Ayhan, Philipp Berens, and
  Siegfried Wahl.
\newblock Leveraging uncertainty information from deep neural networks for
  disease detection.
\newblock \emph{Sci. Rep.}, 7\penalty0 (1):\penalty0 17816, December 2017.

\bibitem[Lipton et~al.(2018)Lipton, Wang, and Smola]{lipton2018detecting}
Zachary Lipton, Yu-Xiang Wang, and Alexander Smola.
\newblock Detecting and correcting for label shift with black box predictors.
\newblock In \emph{International conference on machine learning}, pages
  3122--3130. PMLR, 2018.

\bibitem[Lu et~al.(2018)Lu, Liu, Dong, Gu, Gama, and Zhang]{lu2018learning}
Jie Lu, Anjin Liu, Fan Dong, Feng Gu, Joao Gama, and Guangquan Zhang.
\newblock Learning under concept drift: A review.
\newblock \emph{IEEE Transactions on Knowledge and Data Engineering},
  31\penalty0 (12):\penalty0 2346--2363, 2018.

\bibitem[Maas et~al.(2011)Maas, Daly, Pham, Huang, Ng, and Potts]{imdb}
A.~L. Maas, R.~E. Daly, P.~T. Pham, D.~Huang, A.~Y. Ng, and C.~Potts.
\newblock Learning word vectors for sentiment analysis.
\newblock \emph{The 49th Annual Meeting of the Association for Computational
  Linguistics (ACL).}, 2011.

\bibitem[Pietraszek(2005)]{Pietraszek2005-df}
Tadeusz Pietraszek.
\newblock Optimizing abstaining classifiers using {ROC} analysis.
\newblock In \emph{Proceedings of the 22nd international conference on Machine
  learning}, pages 665--672, 2005.

\bibitem[Platt(1999)]{platt}
J.~C. Platt.
\newblock Probabilistic outputs for support vector machines and comparisons to
  regularized likelihood methods.
\newblock In \emph{Advances in Large Margin Classifiers}, pages 61--74. MIT
  Press, 1999.

\bibitem[Saerens et~al.(2002)Saerens, Latinne, and
  Decaestecker]{Saerens2002-jh}
Marco Saerens, Patrice Latinne, and Christine Decaestecker.
\newblock Adjusting the outputs of a classifier to new a priori probabilities:
  a simple procedure.
\newblock \emph{Neural Comput.}, 14\penalty0 (1):\penalty0 21--41, January
  2002.

\bibitem[Schoelkopf et~al.(2012)Schoelkopf, Janzing, Peters, Sgouritsa, Zhang,
  and Mooij]{Schoelkopf2012-px}
Bernhard Schoelkopf, Dominik Janzing, Jonas Peters, Eleni Sgouritsa, Kun Zhang,
  and Joris Mooij.
\newblock On causal and anticausal learning.
\newblock June 2012.

\bibitem[Srivastava et~al.(2014)Srivastava, Hinton, Krizhevsky, Sutskever, and
  Salakhutdinov]{dropoutsrivastava}
Nitish Srivastava, Geoffrey Hinton, Alex Krizhevsky, Ilya Sutskever, and Ruslan
  Salakhutdinov.
\newblock Dropout: A simple way to prevent neural networks from overfitting.
\newblock \emph{J. Mach. Learn. Res.}, 15:\penalty0 1929--1958, 2014.

\bibitem[Storkey(2009)]{storkey2009training}
Amos Storkey.
\newblock When training and test sets are different: characterizing learning
  transfer.
\newblock \emph{Dataset shift in machine learning}, 30:\penalty0 3--28, 2009.

\bibitem[Sugiyama et~al.(2007)Sugiyama, Krauledat, and
  M{\"u}ller]{sugiyama2007covariate}
Masashi Sugiyama, Matthias Krauledat, and Klaus-Robert M{\"u}ller.
\newblock Covariate shift adaptation by importance weighted cross validation.
\newblock \emph{Journal of Machine Learning Research}, 8\penalty0 (5), 2007.

\bibitem[Wan(1990)]{entropyabstention1}
E~A Wan.
\newblock Neural network classification: a bayesian interpretation.
\newblock \emph{IEEE Trans. Neural Netw.}, 1\penalty0 (4):\penalty0 303--305,
  1990.

\bibitem[Zhu et~al.(2010)Zhu, Gibson, Jun, Rogers, Harrison, and
  Kalish]{zhu2010cognitive}
Xiaojin Zhu, Bryan~R Gibson, Kwang-Sung Jun, Timothy~T Rogers, Joseph Harrison,
  and Chuck Kalish.
\newblock Cognitive models of test-item effects in human category learning.
\newblock In \emph{ICML}, 2010.

\end{thebibliography}
\bibliographystyle{plainnat}

\newpage

\begin{appendix}
\counterwithin{table}{section}
\counterwithin{figure}{section}

\onecolumn
\allowdisplaybreaks

\newpage

\section{Additional Abstention Experiments}
\label{sec:additionalabstentionexperiments}

\subsection{Results for Identifying Pre-Leukemic Hematopoetic Stem Cells}

\begin{table*}[!h]
\tiny
\begin{center}
\begin{tabular}{ | c c | c c c | c c c | }
\hline
\multicolumn{2}{| c |}{ } & \multicolumn{3}{ c |}{
Sensitivity @ 80\% Specificity (pHSCs)} & \multicolumn{3}{c | }{Area under ROC (pHSCs)}\\
\multicolumn{1}{| c}{Method} & \multicolumn{1}{c |}{Calib?} & Base & @30\% Abst. & @15\% Abst. & Base & @30\% Abst. & @15\% Abst.\\ \hline
Est $\Delta$Metric (Ours) & Y & 0.6875$\pm$0.0031 & \textbf{0.8371$\pm$0.0027} & \textbf{0.7638$\pm$0.0027} & 0.8309$\pm$0.0012 & \textbf{0.883$\pm$0.0011} & \textbf{0.8567$\pm$0.0012}\\ \hline
JS Div. (Ours) & Y & 0.6875$\pm$0.0031 & 0.8288$\pm$0.0031 & 0.761$\pm$0.0029 & 0.8309$\pm$0.0012 & 0.8828$\pm$0.0011 & 0.8566$\pm$0.0012\\ \hline
\hline
Dist. from 0.5 & Y & 0.6875$\pm$0.0031 & 0.7484$\pm$0.0021 & 0.712$\pm$0.0028 & 0.8309$\pm$0.0012 & 0.8653$\pm$0.001 & 0.8462$\pm$0.0013\\ \hline
Dist. from 0.5 & N & 0.6875$\pm$0.0031 & 0.7508$\pm$0.0085 & 0.7169$\pm$0.0059 & 0.8309$\pm$0.0012 & 0.8646$\pm$0.0034 & 0.8465$\pm$0.0023\\ \hline
Fumera et al. & Y & 0.6875$\pm$0.0031 & 0.8157$\pm$0.0033 & 0.7355$\pm$0.0033 & 0.8309$\pm$0.0012 & 0.8821$\pm$0.0012 & 0.8553$\pm$0.0012\\ \hline
Fumera et al. & N & 0.6875$\pm$0.0031 & 0.8143$\pm$0.0056 & 0.7376$\pm$0.0052 & 0.8309$\pm$0.0012 & 0.881$\pm$0.0016 & 0.8522$\pm$0.0018\\ \hline
Test-time Dropout & Y & 0.6875$\pm$0.0031 & 0.6816$\pm$0.0108 & 0.6833$\pm$0.0061 & 0.8309$\pm$0.0012 & 0.8318$\pm$0.0047 & 0.8305$\pm$0.0026\\ \hline
Test-time Dropout & N & 0.6875$\pm$0.0031 & 0.6771$\pm$0.0112 & 0.6819$\pm$0.0071 & 0.8309$\pm$0.0012 & 0.8285$\pm$0.0053 & 0.8294$\pm$0.0035\\ \hline
\end{tabular}
\end{center}
\caption{\small \textbf{Sensitivity at 80\% specificity and auROC for identifying regions active in preleukemic Hematopoetic Stem Cells (pHSCs)}. Values represent the mean and standard error across samples generated using a combination of different model initializations and different random splits of the heldout set. ``Base'' indicates performance without abstention. ``Calib?'' indicates whether predicted probabilities were calibrated using Platt scaling \citep{platt}. Within a column, bold values are significantly better than non-bold values according to a Wilcoxon signed rank test (p < 0.05). Ratio of positives:negatives was roughly 1:2. See Sec. ~\ref{sec:genomicsdataset} for more details on the data. Analogous results for genomic regions active in Leukemia Stem Cells (LSCs) are in Table ~\ref{tab:heme_task1}}
\label{tab:heme_task0}
\end{table*}

\FloatBarrier
\clearpage
\subsection{A Note on Deterministic vs. MC Dropout}

In our experiments, for all methods except ``MC Dropout Var.'', the model's predictions were obtained by disabling dropout during test-time - i.e. the predictions were deterministic, and weight rescaling was applied to all layers that used dropout during training \citep{dropoutsrivastava}. However, both the original dropout paper and several subsequent works have noted that it is possible to improve on deterministic dropout by leaving dropout enabled during test-time and taking the expected value of the predictions over a sufficiently large number of Monte Carlo dropout runs \citep{dropoutsrivastava, Gal2015-sn, Leibig2017-ie,  Durr2018-yt}. This is not an inherently surprising result, given that deterministic dropout is intended as a fast approximation of Monte Carlo dropout. Because deterministic dropout is most commonly used in the literature during test-time, we focused our comparisons on this case. However, we also investigated how our proposed abstention framework behaved when predictions were derived by taking the expected value over 100 MC dropout runs, and found that it still performed the best (Supp. Tables ~\ref{tab:dr_noweightrescale_nolabelshift} \& ~\ref{tab:dr_noweightrescale_yeslabelshift}).

\begin{table*}[!h]
\tiny
\begin{center}
\begin{tabular}{ | c c | c c c c c c | }
\hline
\multicolumn{2}{| c |}{ } & \multicolumn{6}{ c |}{Diabetic Retinopathy Cohen's Weighted Kappa (no label shift)}\\
\multicolumn{1}{| c}{Method} & \multicolumn{1}{c |}{Calib?} & Base & @30\% Abst. & @25\% Abst. & @20\% Abst. & @15\% Abst. & @10\% Abst. \\ \hline
Est $\Delta$Metric (Ours) & Y & 0.8106$\pm$0.0018 & \textbf{0.8948$\pm$0.0013} & \textbf{0.8905$\pm$0.0014} & \textbf{0.8842$\pm$0.0014} & \textbf{0.8739$\pm$0.0015} & \textbf{0.8619$\pm$0.0016}\\ \hline
JS Div. (Ours) & Y & 0.8106$\pm$0.0018 & 0.8555$\pm$0.0014 & 0.8521$\pm$0.0014 & 0.8488$\pm$0.0014 & 0.8435$\pm$0.0014 & 0.8369$\pm$0.0015\\ \hline
\hline
Entropy & Y & 0.8106$\pm$0.0018 & 0.7407$\pm$0.0078 & 0.7949$\pm$0.005 & 0.8284$\pm$0.0033 & 0.841$\pm$0.0026 & 0.8401$\pm$0.0021\\ \hline
Entropy & N & 0.8106$\pm$0.0018 & 0.7351$\pm$0.0076 & 0.7789$\pm$0.0052 & 0.812$\pm$0.0037 & 0.8351$\pm$0.0027 & 0.838$\pm$0.0021\\ \hline
Max Class Prob. & Y & 0.8106$\pm$0.0018 & 0.7348$\pm$0.0076 & 0.776$\pm$0.0053 & 0.8054$\pm$0.0037 & 0.8222$\pm$0.0029 & 0.826$\pm$0.0024\\ \hline
Max Class Prob. & N & 0.8106$\pm$0.0018 & 0.7361$\pm$0.0075 & 0.774$\pm$0.0052 & 0.8024$\pm$0.0037 & 0.8174$\pm$0.0029 & 0.8236$\pm$0.0025\\ \hline
Test-time Dropout & Y & 0.8106$\pm$0.0018 & 0.6613$\pm$0.0102 & 0.7086$\pm$0.0069 & 0.7443$\pm$0.0052 & 0.7714$\pm$0.0037 & 0.7844$\pm$0.0031\\ \hline
Test-time Dropout & N & 0.8106$\pm$0.0018 & 0.6775$\pm$0.0101 & 0.73$\pm$0.0067 & 0.7575$\pm$0.0049 & 0.7772$\pm$0.0038 & 0.7867$\pm$0.003\\ \hline
\end{tabular}
\end{center}
\caption{\small \textbf{Cohen's Weighted Kappa for Diabetic Retinopathy (DR) Detection (no label shift), with predictions derived using the expected value of several Monte Carlo dropout runs}. This table is analogous to Table ~\ref{tab:dr_yesweightrescale_nolabelshift}, but predictions for Est $\Delta$ Metric, JS Div., Entropy and ``Max Class Prob.'' were obtained by taking the expected value over 100 MC Dropout runs, rather than by using deterministic (weight rescaling) dropout. Other aspects of the experiment were the same.}
\label{tab:dr_noweightrescale_nolabelshift}
\end{table*}

\begin{table*}[!h]
\tiny
\begin{center}
\begin{tabular}{ | c c | c c c c c c | }
\hline
\multicolumn{2}{| c |}{ } & \multicolumn{6}{ c |}{Diabetic Retinopathy Cohen's Weighted Kappa (with label shift)}\\
\multicolumn{1}{| c}{Method} & \multicolumn{1}{c |}{Adapted?} & Base & @30\% Abst. & @25\% Abst. & @20\% Abst. & @15\% Abst. & @10\% Abst. \\ \hline
Est $\Delta$Metric (Ours) & Y & 0.7837$\pm$0.0044 & \textbf{0.8604$\pm$0.0027} & \textbf{0.8536$\pm$0.0028} & \textbf{0.8462$\pm$0.0028} & \textbf{0.8363$\pm$0.0031} & \textbf{0.8246$\pm$0.0034}\\ \hline
Est $\Delta$Metric (Ours) & N & 0.7313$\pm$0.004 & 0.8275$\pm$0.0035 & 0.818$\pm$0.0035 & 0.8076$\pm$0.0036 & 0.7975$\pm$0.0037 & 0.7862$\pm$0.0038\\ \hline
JS Div. (Ours) & Y & 0.7837$\pm$0.0044 & 0.8457$\pm$0.0027 & 0.8459$\pm$0.0027 & 0.8417$\pm$0.0028 & 0.8349$\pm$0.003 & \textbf{0.8248$\pm$0.0034}\\ \hline
JS Div. (Ours) & N & 0.7313$\pm$0.004 & 0.8515$\pm$0.0029 & 0.8424$\pm$0.0031 & 0.8313$\pm$0.0031 & 0.8174$\pm$0.0033 & 0.7986$\pm$0.0034\\ \hline
\hline
Entropy & Y & 0.7837$\pm$0.0044 & 0.8464$\pm$0.0045 & 0.8386$\pm$0.0046 & 0.8309$\pm$0.0045 & 0.822$\pm$0.0046 & 0.8115$\pm$0.0048\\ \hline
Entropy & N & 0.7313$\pm$0.004 & 0.7802$\pm$0.0053 & 0.7773$\pm$0.0049 & 0.7741$\pm$0.0048 & 0.7684$\pm$0.0046 & 0.7591$\pm$0.0043\\ \hline
Max Class Prob. & Y & 0.7837$\pm$0.0044 & 0.8372$\pm$0.0052 & 0.8294$\pm$0.0052 & 0.8213$\pm$0.0049 & 0.8137$\pm$0.0048 & 0.8051$\pm$0.0046\\ \hline
Max Class Prob. & N & 0.7313$\pm$0.004 & 0.777$\pm$0.0055 & 0.7704$\pm$0.0052 & 0.7673$\pm$0.0048 & 0.7632$\pm$0.0044 & 0.7535$\pm$0.0042\\ \hline
Test-time Dropout & Y & 0.7837$\pm$0.0044 & 0.7954$\pm$0.0048 & 0.7874$\pm$0.0044 & 0.7819$\pm$0.0043 & 0.7771$\pm$0.0043 & 0.7741$\pm$0.0045\\ \hline
Test-time Dropout & N & 0.7313$\pm$0.004 & 0.7567$\pm$0.0055 & 0.7494$\pm$0.0052 & 0.7445$\pm$0.0049 & 0.7384$\pm$0.0044 & 0.7279$\pm$0.0044\\ \hline
\end{tabular}
\end{center}
\caption{\small \textbf{Cohen's Weighted Kappa for Diabetic Retinopathy (DR) Detection, with label shift and predictions derived using the expected value of several Monte Carlo dropout runs}. This table is analogous to Table ~\ref{tab:dr_yesweightrescale_yeslabelshift}, but predictions for Est $\Delta$ Metric, JS Div., Entropy and ``Max Class Prob.'' were obtained by taking the expected value over 100 MC Dropout runs, rather than by using deterministic (weight rescaling) dropout. Other aspects of the experiment were the same.}
\label{tab:dr_noweightrescale_yeslabelshift}
\end{table*}

\FloatBarrier
\clearpage
\section{Runtime Analysis for Algorithm ~\ref{alg:sensatspec}}
\label{sec:runtime_analysis}

In this section, we show that Algorithm ~\ref{alg:sensatspec} has a runtime of $O(NM)$. To see this, consider the steps inside the for loop over $M$. In step (1), we sample $N$ Bernoulli outcomes which takes $O(N)$. In step (2), $\mathbf{n}^+$ and $\mathbf{n}^-$ can be computed in $O(N)$ time via running sums. In step (3), $t^*$ can be found in $O(N)$ time by iterative search. Step (4) takes $O(d)$ time because at most $d$ negatives can be abstained on. In step (5), we calculate $w^-_i$ and $w^+_i$ for $0 \le i \le N - d$ which takes $O(N)$ time. The loop in step (6) is $O(N)$. The normalization step at the end of the algorithm also takes $O(N)$. Note that the calculation of $t^\leftarrow_j$ and $t^\rightarrow_j$ for $0 \le j \le d$ can be accomplished in $O(d)$ time by starting at index $t^*$ and iteratively decrementing (or incrementing) the index until thresholds that satisfy the desired conditions are found. To reduce the number of Monte Carlo samples needed, the output of Algorithm ~\ref{alg:sensatspec} can be smoothed using a Savitzky-Golay filter. In this work, we used a Savitzky-Golay filter of polynomial order 1 and window size 11, and report results with $M=100$.

\FloatBarrier
\clearpage
\section{Optimization Algorithms for AuROC}
\label{sec:appendixoptimauroc}

This section accompanies the discussion in Sec. ~\ref{sec:optimauroc} of the main text. In subsection ~\ref{sec:appendixauroccode}, we present the pseudocode for the $O(MN)$ Monte-Carlo algorithm as well as the $O(N)$ deterministic algorithm for estimating the auROC after a particular interval of examples is abstained on. We also include a comparison of the output of the two algorithms.

Let $\mathbf{p}$ denote a length $N$ vector of calibrated predicted probabilities sorted in ascending order. Let $y_i$ denote the associated labels, and let $s^-_i$ and $s^+_i$ denote the number of negatives and positives ranked below $i$, respectively (i.e. $s^-_i \coloneqq \sum_{i' < i} 1-y_{i'}$ and $s^+_i \coloneqq \sum_{i' < i} y_{i'}$). Let $n^-$ and $n^+$ denote the total number of negatives and positives respectively, i.e. $n^- \coloneqq \sum_i (1-y_i)$ and $n^+ \coloneqq \sum_i y_i$. If the values of $y_i$ are known, then $s^+_i$, $s^-_i$, $n^+$ and $n^-$ can each be calculated in $O(N)$ time for all $i$ via running sums. The probability that an example at index $i$ is ranked above a random negative is $s^-_i/n^-$. Thus, the auROC, being the probability that a random positive is ranked above a random negative, is $\frac{1}{n^- n^+} \sum_{i} y_i s^-_i $. If we define $S \coloneqq \sum_{i} y_i s^-_i$, we can write the auROC as $\frac{S}{n^- n^+}$. The quantity $S$ can be computed from $s_i^-$ and $y_i$ in $O(N)$ time via a running sum.

Let us consider the case of abstaining on indices in $[I_0, I_0 + d)$. Let $w^-_{I_0}$ and $w^+_{I_0}$ denote the number of negatives and positives in the interval respectively, i.e. $w^-_{I_0} \coloneqq \sum_{I_0 \le i < (I_0 + d)} (1-y_i)$ and $w^+_{I_0} \coloneqq \sum_{I_0 \le i < (I_0 + d)} y_i$. Note that $w^-_{I_0}$ and $w^+_{I_0}$ can each be computed for all $I_0$ in $O(N)$ time via running window sums. Further, let us define $W_{I_0} \coloneqq \sum_{I_0 \le i < (I_0 + d)} y_i s^-_i$, which can also be calculated from $y_i$ and $s^-_i$ in $O(N)$ time via a running window sum. Let $s^{I_0,-}_i$ represent the value of $s^-_i$ after abstention in the interval $[I_0, I_0 + d)$, i.e. $s^{I_0,-}_i \coloneqq \mathbbm{1}\{i < I_0\} s^-_i + \mathbbm{1}\{i \ge (I_0+d)\} (s^-_i - w^-_{I_0})$. Note that $s^{I_0,-}_i$ is defined as $0$ for $i \in [I_0, I_0+d)$. Let us introduce the quantity $S^{I_0} \coloneqq \sum_i y_i s^{I_0,-}_i$. We can write the auROC after abstention on interval the $[I_0, I_0 + d)$ as $\frac{S^{I_0}}{(n^- - w^-_{I_0}) (n^+ - w^+_{I_0})}$.

Observe that $S^{I_0} = S - W_{I_0} - \sum_{i \ge I_0 + d} y_i w_{I_0}^- = S - W_{I_0} - (n^+ - s^+_{I_0 + d})w^-_{I_0}$. Thus, $S^{I_0}$ (and therefore the auROC after the interval $[I_0, I_0+d)$ is abstained on) can be computed for all $I_0$ in $O(N)$ time given the values of $S$, $W_{I_0}$, $n^+$, $s^+_i$ and $w^-_i$. The challenge, of course, is what to use for the labels $y_i$, given that we do not have access to this at test time. We could estimate $y_i$ via Monte-Carlo sampling, which would result in a $O(MN)$ algorithm where $M$ is the number of Monte-Carlo samples. However, we were able to bypass the sampling by substituting the expected value $p_i$ for $y_i$ in all the formulas, resulting in a deterministic $O(N)$ algorithm. Empirically, we found that the $O(MN)$ algorithm had very high Pearson and Spearman correlation to the $O(N)$ algorithm (Fig. ~\ref{fig:spearmancorr} \&  ~\ref{fig:pearsoncorr}). Thus, in our experiments, we used the $O(N)$ algorithm.

\clearpage
\subsection{Pseudocode for Monte-Carlo and Deterministic Algorithms}
\label{sec:appendixauroccode}
We first present the non-deterministic $O(MN)$ algorithm that relies on Monte-Carlo sampling.

\begin{algorithm}[H]
\SetKwInOut{Input}{input}\SetKwInOut{Output}{output}
 \Input{Number of examples $d$ to abstain on, sorted calibrated vector $\mathbf{p}$ of length $N$}\;
 
 \Output{Vector $\mathbf{o}$ where $o_i$ is the estimated auROC after indices $[i,i+d)$ are abstained on}\;
 
 Initialize $\mathbf{o}$ to a vector of zeros of length $N + 1 - d$\;
 
 \For{$m \leftarrow 1$ \KwTo $M$}{
    Sample the vector of labels $\mathbf{y}$ according to the probabilities $\mathbf{p}$\;
    
    Compute $n^+ \coloneqq \sum_i y_i$ and $n^- \coloneqq \sum_i (1-y_i)$. This takes $O(N)$ time.\;
    
    Compute $s_i^- \coloneqq \sum_{i' < i} 1 - y_{i'}$ and $s_+^- \coloneqq \sum_{i' < i} y_{i'}$ for all $i$ in $O(N)$ time using running sums.\;
    
    Compute $S \coloneqq \sum_i y_i s^-_i$. This takes $O(N)$ time.\;
    
    Compute $w_i^- \coloneqq \sum_{i \le i' < i+d} (1-y_{i'})$ and $w_i^+ \coloneqq \sum_{i \le i' < i+d} y_{i'}$ for all $0 \le i \le (N-d)$ in $O(N)$ time using running window sums.\;
    
    Compute $W_{i} \coloneqq \sum_{i \le i' < i+d} y_{i'} s_{i'}^-$ for all $0 \le i \le (N-d)$ in $O(N)$ time using a running window sum.\;
    
    Compute $S^i = S - W_i - (n^+ - s^+_{i+d})w^-_i$ for all $0 \le i \le N-d$. This takes $O(N)$ time.\;
    
    Update $\mathbf{o}$ with $o_i \leftarrow o_i + \frac{S^i}{(n^- - w^-_i)(n^+ - w^+_i)}$ for all $0 \le i \le N-d$. This takes $O(N)$ time.\;
 }
 Normalize $\mathbf{o}$ by $M$ using $o_i \leftarrow o_i/M$. This takes $O(N)$ time.\;
 
 Optionally smooth $\mathbf{o}$ using a Savitzky-Golay filter.
 \caption{\small $O(MN)$ Monte-Carlo Algorithm for Optimizing Area under the ROC Curve}
 \label{alg:montecarloauroc}
\end{algorithm}

The deterministic $O(N)$ algorithm is very similar to the Monte-Carlo algorithm, with the key difference being that $p_i$ is used everywhere instead of the sampled label $y_i$.

\begin{algorithm}[H]
\SetKwInOut{Input}{input}\SetKwInOut{Output}{output}
 \Input{Number of examples $d$ to abstain on, sorted calibrated vector $\mathbf{p}$ of length $N$}\;
 
 \Output{Vector $\mathbf{o}$ where $o_i$ is the estimated auROC after indices $[i,i+d)$ are abstained on}\;

Compute $\hat{n}^+ \coloneqq \sum_i p_i$ and $\hat{n}^- \coloneqq \sum_i (1-p_i)$. This takes $O(N)$ time.\;

Compute $\hat{s}_i^- \coloneqq \sum_{i' < i} 1 - p_{i'}$ and $\hat{s}_+^- \coloneqq \sum_{i' < i} p_{i'}$ for all $i$ in $O(N)$ time using running sums.\;

Compute $\hat{S} \coloneqq \sum_i p_i \hat{s}^-_i$. This takes $O(N)$ time.\;

Compute $\hat{w}_i^- \coloneqq \sum_{i \le i' < i+d} (1-p_{i'})$ and $\hat{w}_i^+ \coloneqq \sum_{i \le i' < i+d} p_{i'}$ for all $0 \le i \le (N-d)$ in $O(N)$ time using running window sums.\;

Compute $\hat{W}_{i} \coloneqq \sum_{i \le i' < i+d} p_{i'} \hat{s}_{i'}^-$ for all $0 \le i \le (N-d)$ in $O(N)$ time using a running window sum.\;

Compute $\hat{S}^i = \hat{S} - \hat{W}_i - (\hat{n}^+ - \hat{s}^+_{i+d})\hat{w}^-_i$ for all $0 \le i \le N-d$. This takes $O(N)$ time.\;

Compute $o_i = \frac{\hat{S}^i}{(\hat{n}^- - \hat{w}^-_i)(\hat{n}^+ - \hat{w}^+_i)}$ for all $0 \le i \le N-d$. This takes $O(N)$ time.\;

 \caption{$O(N)$ Deterministic Algorithm for Optimizing Area under the ROC Curve}
 \label{alg:deterministicauroc}
\end{algorithm}

\clearpage
\subsection{Comparison of \texorpdfstring{$O(N)$} and \texorpdfstring{$O(MN)$} algorithms}

\begin{figure}[!h]
\begin{center}
\centerline{\includegraphics[width=0.5\columnwidth]{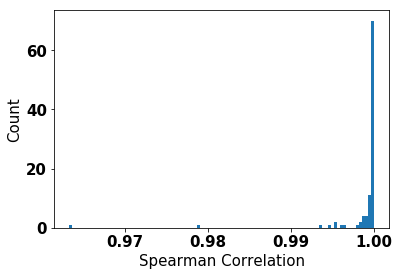}}
\caption{\small \textbf{Histogram of Spearman Correlation between outputs of $O(MN)$ and $O(N)$ auROC optimization algorithms over 100 experiments}.}
\label{fig:spearmancorr}
\end{center}
\end{figure}

\begin{figure}[!h]
\begin{center}
\centerline{\includegraphics[width=0.5\columnwidth]{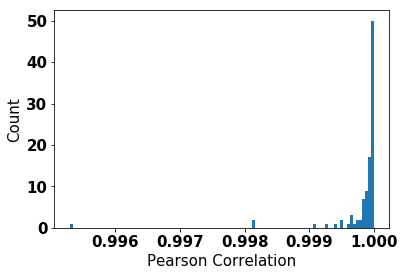}}
\caption{\small \textbf{Histogram of Pearson Correlation between $O(MN)$ outputs of and $O(N)$ auROC optimization algorithms over 100 experiments}.}
\label{fig:pearsoncorr}
\end{center}
\end{figure}

To compare the $O(N)$ and $O(MN)$ algorithms for optimizing auROC, we returned to the binary simulation described in Sec. ~\ref{sec:simulatedbinarydataset}. We generated 100 simulated datasets, each containing $n=1000$ examples, where the parameters for each simulation were sampled uniformly at random from the intervals $q \in [0.1, 0.9)$, $\mu^+ \in [0,5)$, $\mu^- \in [\mu^+ - 5, \mu^+)$, $\sigma^+ \in [1,5)$ and $\sigma^- \in [1,5)$. For each simulated dataset, we ran Algorithms ~\ref{alg:montecarloauroc} and ~\ref{alg:deterministicauroc} with $d=100$ and computed the Spearman and Pearson correlation between the resulting abstention scores. For Algorithm ~\ref{alg:montecarloauroc}, we used $M=1000$ and smoothed the output using a Savitzky-Golay filter of polynomial order 1 and window size 11. Histograms of the Spearman and Pearson correlations over the 100 simulated datasets are in Figures ~\ref{fig:spearmancorr} \& ~\ref{fig:pearsoncorr}. The lowest observed Spearman correlation was greater than $0.96$, and the lowest observed Pearson correlation was greater than $0.996$. In both cases, the mode was very close to 1.0.

\FloatBarrier

\clearpage
\section{Optimization Algorithm for Weighted Cohen's Kappa}
\label{sec:appendixoptimcohenskappa}

\subsection{Derivation of abstention algorithm}

This section continues the discussion in Sec. ~\ref{sec:optimcohenskappa}. As a reminder, Cohen's Kappa applies to the multiclass setting where a predictor $f$ outputs a class $f(x)$ given an example $x$. Let $S$ denote the set of all examples, $N$ denote the size of $S$, $y(x)$ denote the true class of example $x$, and $W_{i,j}$ denote the penalty for predicting an example from class $i$ as being of class $j$. Let $N^i = \sum_{x \in S} \mathbbm{1}\{ y(x) = i\}$ denote the true number of examples in class $i$, and $F^i \coloneqq \sum_S \mathbbm{1}\{f(x) = i\}$ denote the number of examples predicted by $f$ as as being in class $i$. The weighted Cohen's Kappa metric is defined as

\begin{align*}
    \kappa(S,f) \coloneqq 1 - \frac{ \sum_{x \in S} W_{y(x),f(x)} }{\sum_i \sum_j \left( W_{i,j} \frac{N^i}{N} F^j \right)  }
\end{align*}

Let $\kappa(x,S,f)$ represent the new value of $\kappa(S,f)$ when example $x$ is abstained on. We have:

\begin{align*}
    &\kappa(x,S,f) = \left(1 - \frac{\left(\sum_{x' \in S} W_{y(x'),f(x')} \right) - W_{y(x),f(x)}}{ \sum_i \sum_j \left( W_{i,j} \frac{N^i - \mathbbm{1}\{i=y(x)\}}{N-1} (F^j - \mathbbm{1}\{ j = f(x)\} )\right)}\right)
\end{align*}

We can separate this into a sum over different possible values of $y(x)$ and $y(x')$ by writing:

\begin{align}
    &\kappa(x,S,f) = \left(1 - \sum_k \mathbbm{1}\{y(x) = k\} \frac{\left( \sum_{x' \in S} \sum_i W_{i,f(x')}  \mathbbm{1}\{y(x') = i\} \right) -  W_{k,f(x)}}{ \sum_i \sum_j \left( W_{i,j} \frac{N^i - \mathbbm{1}\{i=k\}}{N - 1} (F^j - \mathbbm{1}\{ j = f(x)\} ) \right) } \right) \nonumber \\
    \label{eqn:exactdeltakappa}
    &= \left( 1 - \sum_k \mathbbm{1}\{y(x) = k\} \frac{ (\sum_{x' \in S} \sum_i W_{i,f(x')} \mathbbm{1}\{y(x') = i\}) - W_{k, f(x)} }{ \left( \sum_i \sum_j W_{i,j} \frac{N^i}{N-1} F^j \right) - \left( \sum_j \frac{W_{k,j} F^j}{N-1} \right) - \left(  \sum_i \frac{ W_{i,f(x)} N^i }{ N - 1  } \right) + \frac{W_{k,f(x)}}{N - 1} } \right)
\end{align}

Let $C$ denote the total number of classes. If the true class memberships $y(x)$ are known, then $F^i$ and $N^i$ can be computed for all the $C$ possible values of $i$ in $O(NC)$ time, and $\sum_{x' \in S} \sum_i W_{i,f(x')} \mathbbm{1}\{y(x') = i\}$ can be computed in $O(NC)$ time. Given $F^j$ and $N^i$, the quantity $\sum_i \sum_j W_{i,j} \frac{N^i}{N - 1} F^j$ can be computed in $O(C^2)$ time. Similarly, given $F^j$ the quantity $\sum_j \frac{W_{k,j} F^j}{N-1}$ can be computed for all $C$ possible values of $k$ in $O(C^2)$ time, and given $N^i$ the quantity $\sum_i \frac{W_{i,f(x)} N^i }{N - 1}$ can be computed for all $C$ possible values of $f(x)$ in $O(C^2)$ time. After these quantities have been computed, the value of $\kappa(x,S,f)$ can be found for all $N$ examples $x$ in $O(NC)$ time. Putting it all together, we have that the calculation of $\kappa(x,S,f)$ given the true class memberships $y(x)$ takes $O(NC + C^2)$ time. As $C$ is typically small (e.g. 5 for diabetic retinopathy detection), we can assume that $C < N$, giving a runtime of $O(NC)$.

The challenge now is to determine what to use for the true class memberships $y(x)$, as the class labels are not available at test-time. As before, the class labels could be estimated via Monte-Carlo sampling, which would result in an $O(MNC)$ algorithm where $M$ is the number of Monte-Carlo samples. However, it turns out that we can bypass Monte-Carlo sampling by substituting the calibrated class probabilities for $\mathbbm{1}\{y(x) = i\}$, which results in a deterministic $O(NC)$ algorithm. The pseudocode for both the Monte-Carlo and deterministic algorithms is provided below, and their empirical convergence is explored in Fig. ~\ref{fig:cohenskappaconvergence}.

\begin{algorithm}[H]
\SetKwInOut{Input}{input}\SetKwInOut{Output}{output}
 \Input{A penality weight matrix $\mathbf{W} \in \mathbbm{R}^{C \times C}$ and a prediction matrix $\mathbf{P} \in \mathbbm{R}^{N \times C}$, where $P_{x,i}$ is the calibrated probability that example $x$ belongs to class $i$. $N$ is the total number of examples, and $C$ is the total number of classes}\;
 \Output{Output vector $\mathbf{o}$ where $o_x$ denotes the estimated value of the Weighted Cohen's Kappa after example $x$ is abstained on}\;
 
 Initialize $\mathbf{o}$ to a vector of zeros of length $N$\;
 
 \For{$m \leftarrow 1$ \KwTo $M$}{
    Sample the vector of labels $\mathbf{y}$ according to the probabilities $\mathbf{P}$.\;

    For all classes $i$, compute $N^i \coloneqq \sum_x \mathbbm{1}\{y_x = i\}$. This takes $O(NC)$ time.\;
    
    Compute the predicted labels $f_x = \argmax_i P_{x,i}$. This takes $O(NC)$ time.\;
    
    For all classes $i$, compute $F^i \coloneqq \sum_x \mathbbm{1}\{ f_x = i\}$. This takes $O(NC)$ time.\;
    
    Compute $a \coloneqq \sum_x W_{y_x,f_x}$. This takes $O(N)$ time.\;
    
    Compute $b^1 \coloneqq \sum_i \sum_j W_{i,j} \frac{N^i}{N - 1}F^j$. This takes $O(C^2)$ time.\;
    
    For all classes $i$, compute $b^2_i \coloneqq \sum_j \frac{W_{i,j} F^j}{N - 1}$. This takes $O(C^2)$ time.\;
    
    For all classes $i$, compute $b^3_i \coloneqq \sum_j \frac{W_{j,i} N^j}{N - 1}$. This takes $O(C^2)$ time.\;
    
    Update $\mathbf{o}$ with $o_x \leftarrow o_x + \left(1 - \frac{a - W_{y_x, f_x}}{ b^1 - b^2_{y_x} - b^3_{f_x} + \frac{W_{y_x, f_x}}{N-1}}\right)$. This takes $O(N)$ time.\;
 }
 Normalize $\mathbf{o}$ by $M$ using $o_x \leftarrow o_x/M$. This takes $O(N)$ time.\;
 \caption{\small $O(MNC)$ Monte-Carlo Algorithm for Optimizing Weighted Cohen's Kappa}
 \label{alg:montecarlocohenskappa}
\end{algorithm}

\begin{algorithm}[H]
\SetKwInOut{Input}{input}\SetKwInOut{Output}{output}
 \Input{A penality weight matrix $\mathbf{W} \in \mathbbm{R}^{C \times C}$ and a prediction matrix $\mathbf{P} \in \mathbbm{R}^{N \times C}$, where $P_{x,i}$ is the calibrated probability that example $x$ belongs to class $i$. $N$ is the total number of examples, and $C$ is the total number of classes}\;
 \Output{Output vector $\mathbf{o}$ where $o_x$ denotes the estimated value of the Weighted Cohen's Kappa after example $x$ is abstained on}\;

For all classes $i$, compute $\hat{N}^i \coloneqq \sum_x P_{x,i}$. This takes $O(NC)$ time.\;

Compute the predicted labels $f_x = \argmax_i P_{x,i}$. This takes $O(NC)$ time.\;

For all classes $i$, compute $F^i \coloneqq \sum_x \mathbbm{1}\{ f_x = i\}$. This takes $O(NC)$ time.\;

Compute $\hat{a} \coloneqq \sum_x \sum_i W_{i,f_x} P_{x,i}$. This takes $O(NC)$ time.\;

Compute $\hat{b}^1 \coloneqq \sum_i \sum_j W_{i,j} \frac{\hat{N}^i}{N - 1}F^j$. This takes $O(C^2)$ time.\;

For all classes $i$, compute $b^2_i \coloneqq \sum_j \frac{W_{i,j} F^j}{N - 1}$. This takes $O(C^2)$ time.\;

For all classes $i$, compute $\hat{b}^3_i \coloneqq \sum_j \frac{W_{j,i} \hat{N}^j}{N - 1}$. This takes $O(C^2)$ time.\;

For all $x$, compute $o_x \coloneqq \sum_i P_{x,i} \left(1 - \frac{\hat{a} - W_{i, f_x}}{ \hat{b}^1 - b^2_{i} - \hat{b}^3_{f_x} + \frac{W_{i,f_x}}{N-1} }\right)$. This takes $O(NC)$ time.\;

 \caption{\small $O(NC)$ Deterministic algorithm for Optimizing Weighted Cohen's Kappa}
 \label{alg:deterministiccohenskappa}
\end{algorithm}

\FloatBarrier
\clearpage
\subsection{Empirical Convergence}

\begin{figure}[h]
\begin{center}
\centerline{\includegraphics[width=0.5\columnwidth]{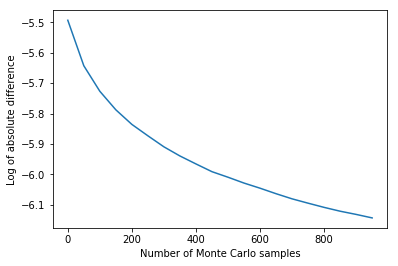}}
\caption{\textbf{Outputs of the $O(MNC)$ Monte-Carlo algorithm and the $O(NC)$ deterministic algorithm converge as the number of Monte Carlo samples is increased}. The log of the mean absolute difference between the output vectors of Algorithm ~\ref{alg:montecarlocohenskappa} and Algorithm ~\ref{alg:deterministiccohenskappa} was plotted as a function of the number of Monte Carlo samples $M$. As $M$ increases, the difference between the outputs appears to monotonically decrease. See the text for details on the simulated dataset used to generate the plot.}
\label{fig:cohenskappaconvergence}
\end{center}
\end{figure}

We explored the empirical convergence of Algorithms ~\ref{alg:montecarlocohenskappa} and ~\ref{alg:deterministiccohenskappa} using a simulated dataset. The simulation was performed as follows: the true class $y_x$ of example $x$ was determined by sampling from a categorical distribution with prior class probabilities of $p(0)=0.4$, $p(1)=0.3$, $p(2)=0.2$ and $p(3)=0.1$. After sampling $y_x$, a value $v_x$ was drawn from a normal distribution with a mean of $\mu_{y_x}$ and a standard deviation of $\sigma_{y_x}$. The means were set to $\mu_0 = -8$, $\mu_1 = -3$, $\mu_2 = 3$, $\mu_3 = 4$ and the standard deviations were set to $\sigma_0 = 4$, $\sigma_1 = 3$, $\sigma_2 = 3$, $\sigma_3 = 2$. Once $v_x$ was sampled, the calibrated posterior probability that the example originated from class $i$ given the observed value $v_x$ was calculated analytically according to the data generating distribution. A total of 10,000 examples were sampled. Algorithms ~\ref{alg:montecarlocohenskappa} and ~\ref{alg:deterministiccohenskappa} were run on the generated data, and the mean absolute difference in the resulting output vectors was plotted as a function of the number of Monte Carlo samples $M$ used in Algorithm ~\ref{alg:montecarlocohenskappa}. The result is shown in Fig. ~\ref{fig:cohenskappaconvergence}. As the number of Monte Carlo samples increases, the mean absolute difference in the output vectors steadily decreases. A Colab notebook reproducing the results in available at \url{https://github.com/blindauth/abstention_experiments/blob/master/convergence_experiments/CohensKappaConvergence.ipynb}.

\FloatBarrier
\newpage

\end{appendix}

\end{document}